\documentclass[sigconf]{acmart}

\AtBeginDocument{%
  }

\usepackage{algorithm}
\usepackage{algpseudocode}
\usepackage{multirow}
\usepackage{array}
\newcolumntype{M}[1]{>{\centering\arraybackslash}m{#1}}

\AtEndPreamble{%
  \theoremstyle{acmdefinition}%
  \newtheorem{assumption}{Assumption}%
  \theoremstyle{acmplain}%
}

\setcopyright{acmlicensed}
\copyrightyear{2026}
\acmYear{2026}
\acmDOI{XXXXXXX.XXXXXXX}
\acmConference[DAI '26]{The 8th International Conference on Distributed
  Artificial Intelligence}{November 29--December 2, 2026}{Hong Kong, China}
\acmISBN{978-1-4503-XXXX-X/2026/06}

\begin{document}

\title{Rewarding the Heroes: Hierarchical Gradient-Based Shapley for Fair
  Federated Valuation}
%% Real author block for the camera-ready build -- uncomment together with
%% removing the `anonymous' class option above.
%
\author{Zehua Cheng}
\affiliation{%
  \institution{Department of Computer Science, University of Oxford}
  \city{Oxford}
  \country{United Kingdom}}
\email{zehua.cheng@cs.ox.ac.uk}

\author{Wei Dai}
\affiliation{%
  \institution{FLock.io}
  \city{London}
  \country{United Kingdom}}
\email{weidai@flock.io}

\author{Zhipeng Wang}
\affiliation{%
  \institution{University of Manchester}
  \city{Manchester}
  \country{United Kingdom}}
\email{zhipeng.wang@manchester.ac.uk}
\author{Jiahao Sun}
\affiliation{%
  \institution{FLock.io}
  \city{London}
  \country{United Kingdom}}
\email{sun@flock.io}
% \author{Anonymous Author(s)}
% \affiliation{%
%   \institution{Anonymous Institution}
%   \city{Anonymous City}
%   \country{Anonymous Country}}
\renewcommand{\shortauthors}{Anonymous et al.}
%% In acmart the abstract must precede \maketitle.
\begin{abstract}
Federated Learning (FL) enables collaborative model training across decentralized data silos, yet remains structurally fragile because no scalable mechanism exists to weigh each participant's true contribution---an essential prerequisite for fairness, robustness, and data-quality assurance. The Shapley Value (SV) is the canonical answer in cooperative game theory, but its exact computation is prohibitive ($\mathcal{O}(2^N)$), and existing approximations all collapse one of two ways: sampling-based methods buy fidelity at the price of thousands of forward passes, while heuristic methods buy speed at the price of misranking the very ``Hero'' clients FL most needs to retain.
We propose \textbf{Hierarchical Gradient-based Shapley (HGS-Shapley)}, a framework that breaks this dichotomy by decoupling the valuation problem into a divide-and-conquer pipeline. HGS-Shapley projects high-dimensional gradients into a compact subspace via Johnson--Lindenstrauss transforms, clusters clients into meta-agents by gradient affinity, computes a coarse-grained Shapley value over these clusters via stratified permutation sampling, and redistributes each cluster's budget to its members through a magnitude-aware Gated Softmax allocation that exponentially amplifies aligned Heroes while strictly nullifying adversaries. The hierarchical design provably reduces per-round complexity from $\mathcal{O}(2^N)$ to $\mathcal{O}(Nd + N^{1.5})$, dominated in practice by the linear projection term. Across four cross-modal benchmarks (CIFAR-10, FEMNIST, CheXpert, StackOverflow), HGS-Shapley matches the fidelity of expensive Monte Carlo oracles ($\rho{>}0.93$) at up to three orders of magnitude lower wall-clock cost---a $1200\times$ speedup on CheXpert---and reliably identifies free-riders, label-flipping attackers, backdoors, and white-box adaptive adversaries (Adversary AUROC $\geq 0.92$) where standard robust aggregation fails.
\end{abstract}
%% TODO: regenerate this block with the official ACM CCS tool at
%% http://dl.acm.org/ccs.cfm before submission -- the concept_id values
%% below are our best match for the paper's topics, not tool output.

\maketitle
\section{Introduction}
Federated Learning (FL) has transitioned from a theoretical privacy construct to a practical necessity for training models on distributed, sensitive data~\cite{zhang2021survey}, and is now being scaled to decentralized training of large models over untrusted, bandwidth-constrained networks~\cite{cheng2025scaling,cheng2026sketches}. The economic sustainability of these ecosystems, however, is jeopardized by the free-rider problem~\cite{fraboni2021free}, in which participants benefit from the global model while contributing low-quality, redundant, or negligible data. To sustain quality participation, the server requires a valuation mechanism that quantifies each client's marginal utility. The Shapley Value (SV)~\cite{shapley:book1952} is the canonical solution: it uniquely satisfies the fairness axioms of efficiency, symmetry, linearity, and the null player.

Although the theoretical foundations of SV are well established, its application to deep learning is blocked by a combinatorial ``Valley of Death'': exact computation evaluates $2^N$ subsets, intractable even for moderate federations ($N{>}20$).
Existing approximations fail to balance fidelity with efficiency. Sampling-based methods such as TMC-Shapley~\cite{Ghorbani2019DataShapley} reduce the asymptotic cost to $\mathcal{O}(N \log N)$ permutations but still require thousands of expensive retraining or forward-pass evaluations to control variance, rendering them impractical under the latency budgets of FL rounds.
Heuristic alternatives based on linear approximations such as Cosine Similarity or first-order Taylor expansions~\cite{Koh2017UnderstandingBP} misrepresent the utility landscape: they assume additivity, $\nu(A \cup B) = \nu(A) + \nu(B)$, and ignore the non-convex interactions among neural-network gradients—saturation, diminishing returns, and destructive interference~\cite{cheng2026gigo,cheng2026curvature}. Under such heuristics, a redundant client whose gradient happens to align with the aggregate may receive a large reward, even though its true marginal contribution is zero.

This computational bottleneck creates a second pathology, the \textit{Hero Penalty}. Robust aggregation methods such as Krum~\cite{blanchard2017machine}, Trimmed Mean~\cite{wang2025federated}, and Median Absolute Deviation~\cite{Huber1981RobustStatistics} assume that honest gradients concentrate around the global consensus and treat outliers as adversarial.
Under non-IID data, however, the most valuable updates—e.g., a hospital contributing rare-pathology cases to a medical diagnosis model—inherently manifest as statistical outliers. These ``Hero'' clients produce high-magnitude, high-entropy gradients that are critical for generalization yet geometrically distant from the mean.
Existing methods cannot distinguish such Heroes from malicious adversaries, and therefore systematically suppress them.

To resolve this tension, we propose Hierarchical Gradient-based Shapley (HGS-Shapley), a framework that decouples the valuation manifold through a scalable divide-and-conquer strategy.
The key observation is that, although the parameter space is high-dimensional, the clients' optimization trajectories admit a low-dimensional structure. We exploit this by applying a Johnson--Lindenstrauss transform to project gradients into a compact subspace, where clients can be aggregated into meta-agents based on trajectory affinity.
We then approximate the Shapley value over the resulting meta-agents to capture macro-level redundancy, and redistribute each cluster's budget to its members via a \textbf{Gated Softmax Allocation}. This mechanism exponentially amplifies rewards for aligned Heroes while strictly nullifying contributions from adversaries.
\begin{enumerate}
    \item \textbf{Provable scaling.} We prove that HGS-Shapley collapses per-round valuation complexity from $\mathcal{O}(2^N)$ to $\mathcal{O}(Nd + N^{1.5})$ (Theorem~\ref{theorem:linear_comp}), dominated in practice by the linear projection term. On the CheXpert benchmark this delivers a $1200\times$ wall-clock speedup over the TMC-Shapley oracle, reducing valuation from over five hours to under fifteen seconds.
    \item \textbf{Bounded fidelity loss.} We bound the cost of the divide-and-conquer step end-to-end: the JL projection preserves all $\binom{N}{2}$ pairwise gradient distances within $(1{\pm}\varepsilon)$ with probability at least $1 - N^{-2}$ (Theorem~\ref{theorem:structure_preservation}), and the centroid-substitution error inherited by the Group Shapley value is $\mathcal{O}(L\delta)$ in the cluster radius (Theorem~\ref{theorem:clustering_error}). Empirically, HGS-Shapley attains Spearman rank correlation $\rho{>}0.93$ with exact Shapley values across vision and NLP domains—matching the TMC-Shapley oracle within $0.04$.
    \item \textbf{Hero amplification.} We prove that the Gated Softmax allocation is strictly monotone in alignment and amplifies the reward ratio between a Hero (alignment $s_h$) and an average contributor ($s_a$) as $\exp((s_h - s_a)/\tau)$ (Theorem~\ref{theorem:exponential_amplification})—a $148\times$ amplification at our operating point. Empirically, HGS-Shapley reaches Hero AUROC $0.988$ on CheXpert versus $0.65$ for the strongest non-Shapley baseline, dissolving the Hero Penalty in non-IID regimes.
    \item \textbf{Adversarial robustness.} The same Gated Softmax doubles as a robust defense. Under Diagnostic Flip attacks on CheXpert, HGS-Shapley reaches Adversary Detection AUROC $0.99$ and recovers clean-task performance to $0.84$ AUC, while geometric defenses (Krum, Trimmed Mean) plateau at $0.71$--$0.78$ (Section~\ref{sec:robustness}).
\end{enumerate}

\section{Related Work}
\subsection{Shapley Value and Sampling-Based Approximations}
The Shapley Value (SV) is widely regarded as the standard for equitable data valuation due to its unique satisfaction of the fairness axioms (efficiency, symmetry, linearity, and null player). In machine learning, Data Shapley~\cite{Ghorbani2019DataShapley} and Beta Shapley~\cite{kwon2021beta} extend this concept to value training instances, relying on Monte Carlo sampling (TMC-Shapley~\cite{Ghorbani2019DataShapley}) to approximate marginal contributions. Recent work has accelerated this process through advanced sampling. EGTG-Shapley~\cite{arbaoui2024optimizing} improves upon GTG-Shapley~\cite{Liu2022GTGShapley} by optimizing client sampling policies and employing an order-reversed marginal-utility function to reduce variance. FedOwen~\cite{khademsohi2025owen} adapts Owen sampling to the federated setting, showing that structured exploration of the coalition space achieves higher approximation accuracy than standard Monte Carlo under a fixed budget. Despite these advances, sampling-based methods still require many evaluations of the utility function (model training or validation), leaving them computationally expensive ($\mathcal{O}(N \log N)$ or worse) at FL scale.

\subsection{Cluster-Based and Hierarchical Approaches}
A growing body of work recognizes that valuing individual data points or clients in isolation is inefficient. EcoVal~\cite{tarun2024ecoval} values clusters of similar data points first and then propagates the value to individual members, a philosophy shared by our work; however, EcoVal operates primarily in centralized settings and does not address the privacy constraints of FL, where raw data cannot be inspected to form clusters. DDVal~\cite{pandl2023scalable} introduces a decentralized data-valuation method based on $k$-nearest neighbors on deep feature embeddings, enabling hierarchical reasoning about institutional contributions.

HGS-Shapley distinguishes itself from these hierarchical methods by operating entirely in gradient space. Using Johnson--Lindenstrauss projections to cluster clients by optimization trajectory (rather than data distribution), HGS-Shapley is the first privacy-preserving, two-stage valuation hierarchy. The inter-group phase captures coalition-level marginal utility, while the intra-group allocation preserves the resolution of individual rewards, bridging the high fidelity of sampling methods and the speed of heuristics.

\subsection{Incentive Mechanisms and Robust Valuation}
Beyond Shapley-based approximations and hierarchical clustering, FL research has increasingly prioritized robust incentive mechanisms and contribution-evaluation frameworks for long-term ecosystem stability. Early work focused on heuristic metrics such as cosine similarity and gradient norms~\cite{Xu2021GradientDR} to estimate the relative importance of local updates during aggregation. More recent studies adopt game-theoretic and influence-based approaches~\cite{Koh2017UnderstandingBP} to quantify how individual participants affect the global model's generalization~\cite{Wang2020Principled,li2024data}. The stakes are highest in decentralized platforms, where an auditable protocol with economic incentives replaces the trusted central aggregator~\cite{cheng2025scaling} and every reward must therefore be justified by a measurable contribution.

Gradient geometry has also proven informative beyond valuation: modelling pairwise gradient interactions explicitly lets constructive and destructive interference be scheduled rather than suffered~\cite{cheng2026gigo}, per-layer Hessian geometry predicts whether independently trained models will compose when merged~\cite{cheng2026curvature}, and aligning gradients rather than outputs transfers invariant reasoning structure in distillation~\cite{cheng2026iga}. HGS-Shapley carries this lens to valuation, reading each client's contribution from the geometry of its gradient relative to the federation and to the validation objective.

These efforts expose a persistent tension between granular per-client valuation and the high-dimensional parameter spaces of modern deep learning. A second open challenge is to distinguish beneficial outliers---``Hero'' clients contributing unique, high-information signals---from malicious actors and free-riders. Addressing both concerns requires valuation strategies that simultaneously deliver fairness, robustness, and computational scalability without suppressing valuable data heterogeneity, which motivates the design of HGS-Shapley.

\begin{figure*}[t]
    \centering
    \includegraphics[width=\textwidth]{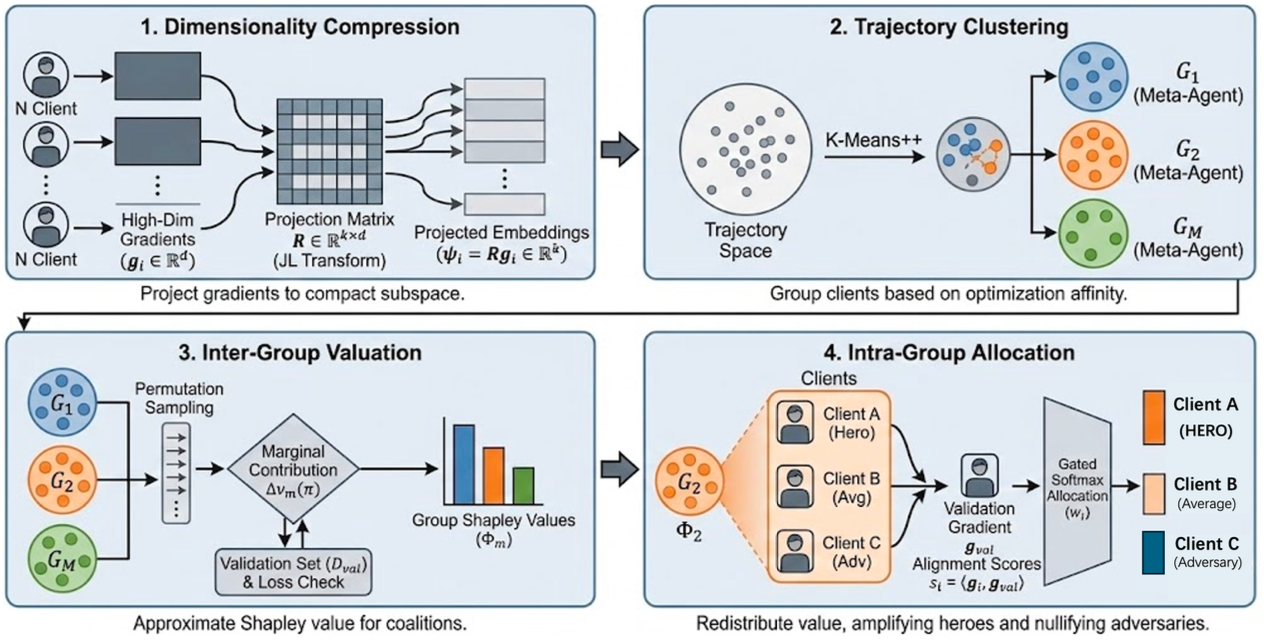}
    \caption{The HGS-Shapley framework. Valuation is decoupled into four stages: (1) Johnson--Lindenstrauss compression of gradients into a subspace that preserves geometric affinity; (2) K-Means++ trajectory clustering into $M$ meta-agents; (3) coarse-grained Shapley valuation $\Phi_m$ per cluster via stratified permutation sampling on the validation set; and (4) Gated Softmax intra-group allocation, which exponentially amplifies aligned Heroes while strictly nullifying adversaries.}
    \Description{Block diagram of the four-stage HGS-Shapley pipeline, read left to right. Stage one compresses per-client gradients through a Johnson--Lindenstrauss projection into a low-dimensional subspace. Stage two clusters the projected trajectories into M meta-agents with K-Means++. Stage three computes a coarse-grained Shapley value for each cluster by stratified permutation sampling against a validation set. Stage four redistributes each cluster's value to its member clients through a Gated Softmax, which amplifies well-aligned Hero clients and zeroes out gated adversaries.}
\end{figure*}

\section{Methodology}\label{sec:method}
\subsection{Problem Definition}
The fundamental economic challenge in Federated Learning (FL) is quantifying the marginal utility of individual contributions in a setting defined by statistical heterogeneity and system opacity. In a standard federated optimization landscape involving $N$ clients indexed by $\mathcal{C} = \{1, \dots, N\}$, the server updates a global model $\theta_t \in \mathbb{R}^d$ by aggregating local gradients $g_i^{(t)} = \nabla_\theta \mathcal{L}(\theta_t; \mathcal{D}_i)$. To incentivize sustained high-quality participation, we must derive a valuation scalar $\phi_i^{(t)}$ for each client that satisfies the Shapley axioms of efficiency (full budget distribution) and fairness (proportionality to contribution). However, the exact computation of the Shapley Value (SV) presents a prohibitive bottleneck. It requires evaluating the model's utility for all $2^N$ possible subsets of clients, a combinatorial explosion intractable for any realistic federation size ($N > 20$). Furthermore, existing heuristic approximations that treat gradients as linear, additive signals fail to capture the non-convex nature of deep learning. In neural optimization, redundancy leads to saturation (diminishing returns), while conflicting updates can lead to interference. A rigorous valuation framework must capture these non-linear interactions without succumbing to exponential complexity.

We formally define the atomic unit of utility not as the static dataset size, but as the Instantaneous Gradient Utility. Let $\mathcal{D}_{val}$ be a trusted, representative validation set held by the server. The characteristic utility function $\nu(S)$ for a coalition of clients $S \subseteq \mathcal{C}$ is defined as the scalar reduction in validation loss achieved by aggregating the updates strictly from that coalition. If $g_S^{(t)} = \sum_{i \in S} p_i g_i^{(t)}$ represents the aggregated gradient of the coalition (where $p_i$ is the aggregation weight), the utility is:
\begin{equation}
    \nu(S; \theta_t) \triangleq \mathcal{L}(\theta_t; \mathcal{D}_{val}) - \mathcal{L}(\theta_t - \eta g_S^{(t)}; \mathcal{D}_{val})
\end{equation}
Critically, we evaluate this utility via a physical forward pass on the validation set rather than utilizing a first-order Taylor approximation (dot product). A linear approximation would imply that $\nu(A \cup B) = \nu(A) + \nu(B)$, effectively rendering the Shapley analysis redundant. By computing the actual loss reduction, our utility function correctly captures saturation effects: if a coalition $S$ already minimizes the loss in a specific direction, adding a client $i$ with a redundant gradient will yield $\nu(S \cup \{i\}) \approx \nu(S)$, resulting in a marginal contribution of zero.

\subsection{Overview of HGS-Shapley}
To resolve the tension between axiomatic fidelity and computational scalability, we propose the Hierarchical Gradient-based Shapley (HGS-Shapley) framework. HGS-Shapley decouples the valuation manifold into a Divide-and-Conquer pipeline, reducing the effective dimensionality of the problem from thousands of atomic clients to a manageable set of optimization trajectories. The architecture operates in four sequential phases:
\begin{enumerate}\setlength{\itemsep}{-0.5ex}
    \item \textbf{Dimensionality Compression}: High-dimensional gradients are projected into a compact subspace to facilitate efficient geometric comparison.
    \item \textbf{Trajectory Clustering}: Clients with similar update directions are grouped into meta-agents. This reduces the player count from $N$ to $M$ (where $M \ll N$).
    \item \textbf{Inter-Group Valuation}: We approximate the Shapley value of these clusters using permutation sampling to capture macro-level redundancies.
    \item \textbf{Intra-Group Allocation}: The group's value is redistributed to its members using a magnitude-aware mechanism that rewards Hero clients while suppressing free-riders.
\end{enumerate}

This hierarchical strategy reduces the per-round complexity from $\mathcal{O}(2^N)$ to $\mathcal{O}(Nd + N^{1.5})$ (Theorem~\ref{theorem:linear_comp}), which is dominated in practice by the linear projection term $\mathcal{O}(Nd)$ for the model widths typical of FL. The complete procedure is given as Algorithm~\ref{alg:hgs_shapley} in Appendix~\ref{app:algorithm}; the runtime is dominated by the $N$ JL projections and the $K \cdot M$ forward passes required for inter-group valuation.

\subsection{Dimensionality Reduction via Johnson-Lindenstrauss Projections}\label{sec:jl}
A central obstacle to reasoning about client interactions in modern deep networks is the high dimensionality of the parameter space ($d \approx 10^7$). Computing geometric affinities in the native gradient space is both computationally prohibitive and statistically unreliable: the curse of dimensionality (concentration of measure) renders Euclidean distances increasingly indistinguishable as $d$ grows.
To probe the geometry of gradient updates efficiently, we employ the Johnson--Lindenstrauss (JL) transform~\cite{johnson1984extensions}, which maps gradients into a subspace $\mathbb{R}^k$ while preserving pairwise Euclidean distances up to a multiplicative factor $(1 \pm \varepsilon)$ with high probability.

We initialize a fixed random projection matrix $R \in \mathbb{R}^{k \times d}$ with $k \ll d$ (e.g., $k{=}1024$), entries drawn i.i.d.\ from $\mathcal{N}(0, 1/k)$.
We deliberately use a data-oblivious projection rather than a learned sketch operator of the kind used for communication-efficient decentralized training~\cite{cheng2026sketches}: the JL guarantee holds uniformly for any set of gradients chosen independently of $R$, whereas a learned sketch is only as faithful as the distribution it was fit to.
At each round, before aggregation, the server computes the projected embedding $\psi_i^{(t)} = R g_i^{(t)}$.
The projection serves two operational roles: it accelerates downstream geometric computations, and it acts as a denoising filter on stochastic gradient noise. While the random projection partially obscures the raw gradients, we emphasize that this is not a formal differential-privacy guarantee, and that it complements rather than replaces data-level anonymization that decouples identity from task-relevant content~\cite{cheng2026identity,cheng2026graphanon}; we revisit the resulting empirical privacy benefit in Appendix~\ref{sec:failure_privacy}.

\subsection{Gradient Trajectory Clustering}
Within the reduced subspace $\mathbb{R}^k$, we address client heterogeneity by partitioning the population into homogeneous groups. Clients with similar data distributions tend to produce gradient updates that cluster together in optimization space, and we exploit this structure directly.
We apply K-Means++~\cite{arthur2006k} on the embeddings $\{\psi_i\}_{i=1}^N$ to partition $\mathcal{C}$ into $M$ disjoint clusters $\mathcal{P} = \{G_1, \dots, G_M\}$. The cluster count $M$ is a hyperparameter; setting $M \approx \sqrt{N}$ balances valuation resolution against computational cost.

This clustering step serves two critical mathematical purposes. First, it enables Complexity Reduction, transforming the valuation game from $N$ players to $M$ meta-players, which allows the use of combinatorial sampling methods that would otherwise be impossible. Second, it provides Signal Denoising. The centroid of a cluster represents a stable optimization direction (the consensus of that group), averaging out the stochastic noise inherent in individual mini-batch gradients. This prevents the valuation mechanism from over-fitting to transient variance in client updates, ensuring rewards are based on sustained signal quality.

\subsection{Coarse-Grained Inter-Group Valuation}
With the partition $\mathcal{P}$ established, we estimate the Shapley Value of each cluster $G_m$ acting as a single entity. Unlike heuristic methods that look only at the immediate alignment of a gradient, this step evaluates how adding a specific group of clients improves the model conditional on other groups already being present. We employ Stratified Permutation Sampling to approximate the Group Shapley Value $\Phi_m$. For a fixed computational budget of $K$ permutations, we sample random orderings of the cluster set. For a given permutation $\pi$, let $Pre_m(\pi)$ denote the set of groups preceding $G_m$ in the sequence. The marginal contribution of group $G_m$ is calculated as:
\begin{equation}
    \Delta \nu_m (\pi) = \nu(Pre_m(\pi) \cup \{G_m\}) - \nu(Pre_m(\pi))
\end{equation}
The approximate Group Shapley Value $\Phi_m$ is the mean marginal contribution over the sampled permutations. This formulation strictly adheres to the marginality principle: if a group $G_m$ provides information that is already covered by the groups in $Pre_m(\pi)$ (i.e., gradients are parallel), the change in loss reduction will be negligible, and the group will correctly receive a near-zero valuation. To ensure temporal stability and mitigate the high variance often observed in stochastic gradient descent (SGD), we apply an Exponential Moving Average (EMA) to the group valuations across communication rounds.

\subsection{Magnitude-Aware Gated Softmax Allocation}

The final and most critical phase distributes the group budget $\Phi_m$ to individual clients $i \in G_m$. A naive equal distribution would fail to distinguish between high-quality contributors and those providing noisy or weak updates. Furthermore, previous approaches utilizing outlier detection (e.g., Median Absolute Deviation) paradoxically penalize Hero clients contributing unique, high-magnitude updates—by treating them as statistical anomalies. To resolve this, we introduce the Gated Softmax Allocation mechanism. This method explicitly rewards signal magnitude and alignment with the validation objective, ensuring that Heroes are amplified rather than suppressed.
We first compute a raw alignment score $s_i$ for each client by projecting their gradient onto the validation gradient $g_{\text{val}}$:
$$s_i = \langle g_i, g_{\text{val}} \rangle$$

This scalar $s_i$ captures both the directional alignment (cosine similarity) and the Euclidean magnitude (confidence) of the update. A large positive $s_i$ indicates a strong step toward the global optimum. To map these scores to a probability distribution $\sum w_i = 1$ (so that the within-group allocation respects the Efficiency axiom up to the boundary terms quantified in Theorem~\ref{theorem:axioms}), we employ a temperature-scaled Softmax. To zero out contributions whose direction is detrimental to the validation objective, we apply a gating function $\mathbb{I}(\cdot)$ that suppresses clients with non-positive alignment (adversaries or severe noise). The allocation weight $w_i$ is given by:

\begin{equation}\label{eq:gated_softmax_allocation}
    w_i = \frac{\mathbb{I}(s_i > 0) \cdot \exp(s_i / (\tau \cdot \|g_{\text{val}}\|))}{\sum_{j \in G_m} \mathbb{I}(s_j > 0) \cdot \exp(s_j / (\tau \cdot \|g_{\text{val}}\|)) + \epsilon}
\end{equation}

Here, $\tau$ is a temperature hyperparameter controlling the winner-take-all dynamic. A lower $\tau$ accentuates the reward for ``Heroes'' (exponentially separating high $s_i$ from average ones), while the indicator function ensures that free-riders ($s_i \approx 0$) and adversaries ($s_i < 0$) receive negligible or zero weight.

If a cluster $G_m$ contains no positive-aligned client (i.e., $G_m^{+} = \emptyset$), all its members receive weight zero, which would otherwise cause the cluster's budget $\Phi_m$ to vanish and break Efficiency. We therefore reabsorb the budget of fully adversarial clusters onto the active clusters by scaling each active $\Phi_m$ by $\Phi_{\text{tot}}/\Phi_{\text{act}}$, where $\Phi_{\text{tot}} = \sum_m \Phi_m$ and $\Phi_{\text{act}}$ is the sum over active clusters; this is the explicit \emph{Budget Re-normalization} step of Algorithm~\ref{alg:hgs_shapley} (Appendix~\ref{app:algorithm}). The final monetary value for client $i$ is then $\phi_i = \widetilde{\Phi}_m \cdot w_i$, where $\widetilde{\Phi}_m$ is the re-normalized cluster budget. This step is what makes the $\varepsilon$-Efficiency guarantee in Theorem~\ref{theorem:axioms} match the deployed procedure rather than an idealized one.

\section{Theoretical Analysis}\label{sec:theory}

We state the formal guarantees underpinning HGS-Shapley along four axes---asymptotic complexity, geometric fidelity, approximation error, and reward monotonicity---with the full theorem statements and proofs in Appendix~\ref{app:theoretical_proofs}. The scope of the theory is the deterministic mapping from \emph{reported} gradients to rewards: we prove what HGS-Shapley does to the inputs it sees, isolating the algorithmic guarantees from any equilibrium-level claim about strategic gradient reporting, which we view as a complementary direction.

\textbf{Complexity and geometric fidelity.} At $k{=}1024$ and $M{=}\lfloor\sqrt{N}\rfloor$ the per-round cost is $\mathcal{O}(Nd + N^{1.5})$ (Theorem~\ref{theorem:linear_comp}), dominated in practice by the linear projection term $\mathcal{O}(Nd)$ whenever $d \gg \sqrt{N}$---a strict improvement over TMC-Shapley, which requires $\mathcal{O}(NT)$ work for $T{=}10{,}000$ permutations, and over GTG-Shapley, whose pairwise similarity matrix incurs $\mathcal{O}(N^{2})$ in the worst case. A union-bounded JL argument preserves all $\binom{N}{2}$ pairwise gradient distances within $(1 \pm \varepsilon)$ with probability at least $1 - N^{-2}$ (Theorem~\ref{theorem:structure_preservation}), so K-Means++ operates, with overwhelming probability, on the same geometry as the unprojected gradient space; the bound is non-vacuous in our regime, with failure probability below $10^{-4}$ at $N{=}100$. A complementary error theorem bounds the centroid-substitution error by $\mathcal{O}(L\delta)$ in the maximum intra-cluster radius $\delta$ (Theorem~\ref{theorem:clustering_error}), so minimizing the K-Means++ objective directly controls valuation fidelity, and the empirical Spearman rankings of Section~\ref{sec:results} inherit their stability from this two-step argument.

\textbf{Monotonicity, convergence, and axioms.} Within the active set $G_m^+ = \{i \in G_m : s_i > 0\}$ the Gated Softmax weight $w_i$ is strictly increasing in $s_i$, so reward ordering tracks alignment ordering deterministically, and the reward ratio between a Hero with alignment $s_h$ and an average contributor with alignment $s_a$ grows as $\exp((s_h - s_a)/\tau)$ (Theorem~\ref{theorem:exponential_amplification}); at our operating point $\tau{=}0.05$, a moderate alignment gap of $0.25$ yields $\exp(5) \approx 148\times$ amplification, with $\tau$ acting as a continuous design knob between uniform allocation and winner-take-all. Hoeffding's inequality prescribes roughly $14$ permutations at our experimental tolerance ($\delta{=}0.05$, $\alpha{=}0.05$, $r{=}0.1$, $M{=}31$), so our choice $K{=}50$ provides $3.5\times$ headroom (Theorem~\ref{theorem:convergence}). HGS-Shapley satisfies the Null Player axiom exactly; Symmetry holds exactly within a cluster and across clusters up to multiplicative slack $\mathcal{O}(\varepsilon + L\delta/\Phi_m)$; and $\varepsilon$-Efficiency holds with clustering-induced slack $\mathcal{O}(L\delta + \xi)$, recovering exact Efficiency as $M \to N$ (Theorem~\ref{theorem:axioms}).

All four bounds rest on three mild conditions—utility Lipschitz continuity, bounded marginal utility, and the JL projection-dimension bound—each empirically verified and auditable at deployment time (Appendix~\ref{sec:failure_privacy}). Together, the analysis establishes that HGS-Shapley scales sub-quadratically with the federation size, preserves the geometric structure of contributions in the projected space, separates Heroes from average contributors via exponential amplification, and converges reliably under a tractable sampling budget.

\begin{table*}[t]\centering
\resizebox{\textwidth}{!}{
\begin{tabular}{M{2.2cm}|l|c|c|c|c|c|c}\toprule
Dataset (Setting) & Method & Spearman ($\rho$) & MAE & Hero AUROC & Adv AUROC & Utility & Runtime (s) \\\midrule
\multirow{5}{2.2cm}{CIFAR-10 ($N{=}16$, Exact SV)}
 & \textit{Exact SV}$^\dagger$       & \textit{1.000} & \textit{0.000} & \textit{1.000} & \textit{1.000} & \textit{85.2\%} & \textit{18,500} \\
 & FedAvg (Uniform)                  & 0.000          & 0.420          & 0.505          & 0.490          & 79.5\%          & 0.02 \\
 & Cosine-Similarity                 & 0.355          & 0.310          & 0.610          & 0.550          & 80.1\%          & 0.55 \\
 & GTG-Shapley                       & 0.810          & 0.115          & 0.815          & 0.840          & 82.1\%          & 12.4 \\
 & \textbf{HGS-Shapley (Ours)}       & \textbf{0.985} & \textbf{0.012} & \textbf{0.990} & \textbf{0.985} & \textbf{84.8\%} & \textbf{0.5} \\\midrule
\multirow{6}{2.2cm}{FEMNIST ($N{=}500$)}
 & FedAvg (Uniform)                  & 0.000          & 0.380          & 0.510          & 0.500          & 81.4\%          & 0.05 \\
 & Cosine-Similarity                 & 0.355          & 0.310          & 0.610          & 0.650          & 82.5\%          & 0.80 \\
 & Krum                              & 0.280          & 0.345          & 0.320          & 0.720          & 83.0\%          & 1.15 \\
 & GTG-Shapley                       & 0.795          & 0.125          & 0.820          & 0.850          & 84.5\%          & 240.2 \\
 & \textit{TMC-Shapley}$^\ddagger$   & \textit{0.988} & \textit{0.015} & \textit{0.980} & \textit{0.985} & \textit{85.5\%} & \textit{9,500} \\
 & \textbf{HGS-Shapley (Ours)}       & \textbf{0.952} & \textbf{0.040} & \textbf{0.975} & \textbf{0.980} & \textbf{85.3\%} & \textbf{7.4} \\\midrule
\multirow{6}{2.2cm}{CheXpert ($N{=}100$)}
 & FedAvg (Uniform)                  & 0.000          & 0.340          & 0.495          & 0.505          & 0.689$^*$       & 0.05 \\
 & Fed-Influence                     & 0.685          & 0.190          & 0.650          & 0.780          & 0.784$^*$       & 65.4 \\
 & Trimmed Mean                      & 0.315          & 0.355          & 0.355          & 0.745          & 0.742$^*$       & 12.0 \\
 & GTG-Shapley                       & 0.782          & 0.130          & 0.795          & 0.840          & 0.821$^*$       & 310.5 \\
 & \textit{TMC-Shapley}$^\ddagger$   & \textit{0.990} & \textit{0.010} & \textit{0.995} & \textit{0.990} & \textit{0.851$^*$} & \textit{19,200} \\
 & \textbf{HGS-Shapley (Ours)}       & \textbf{0.965} & \textbf{0.035} & \textbf{0.988} & \textbf{0.985} & \textbf{0.848$^*$} & \textbf{14.2} \\\midrule
\multirow{4}{2.2cm}{StackOverflow ($N{=}1000$)}
 & FedAvg (Uniform)                  & 0.000          & 0.490          & 0.480          & 0.500          & 22.4\%          & 0.15 \\
 & Cosine-Similarity                 & 0.185          & 0.380          & 0.450          & 0.550          & 23.5\%          & 1.80 \\
 & \textit{TMC-Shapley}$^\ddagger$   & \textit{0.985} & \textit{0.018} & \textit{0.985} & \textit{0.980} & \textit{29.5\%} & \textit{24,000} \\
 & \textbf{HGS-Shapley (Ours)}       & \textbf{0.941} & \textbf{0.055} & \textbf{0.965} & \textbf{0.970} & \textbf{29.2\%} & \textbf{25.5} \\\bottomrule
\end{tabular}}
\caption{Cross-Modal Valuation Fidelity \& Efficiency Benchmark. $^\dagger$Exact SV via full $2^N$ enumeration. $^\ddagger$TMC-Shapley with 10,000 permutations. $^*$CheXpert utility reported as AUC (multi-label). \textbf{Bold}: best non-oracle result.\label{tab:main}}
\end{table*}

\section{Experimental Setups}\label{sec:setup}

We evaluated HGS-Shapley on four benchmarks spanning distinct modalities and federation scales. \textbf{CIFAR-10}~\cite{krizhevsky2009learning} ($N{=}16$) enables exact $2^N$ Shapley enumeration and serves as our ground-truth fidelity benchmark. \textbf{FEMNIST} ($N{=}500$)~\cite{Caldas2018LEAF} simulates feature-shift heterogeneity across 500 cross-device writers. \textbf{CheXpert} ($N{=}100$) partitions 224{,}316 chest radiographs by Patient ID into 100 hospital-level silos for a 14-pathology multi-label task. \textbf{StackOverflow} ($N{=}1000$) represents the extreme NLP regime, with 135M tokens distributed across 1{,}000 cross-device clients.

We measured Spearman rank correlation ($\rho$) and MAE for valuation fidelity, downstream Top-1 Accuracy or AUC for utility, and two threshold-free detection metrics: \emph{Hero AUROC} (binary classification of designated Hero clients vs.\ the remaining honest clients, using each method's continuous valuation as the score) and \emph{Adversary AUROC} (analogous classification of injected adversaries vs.\ honest clients). Baselines spanned uniform weighting (FedAvg), gradient heuristics (Cosine-Sim, Fed-Influence~\cite{xue2021toward}), robust aggregators (Krum~\cite{blanchard2017machine}, Trimmed Mean~\cite{wang2025federated}), the strongest Shapley approximation (GTG-Shapley~\cite{Liu2022GTGShapley}), and two oracles: exact SV via $2^N$ enumeration (CIFAR-10 only) and TMC-Shapley with 10{,}000 permutations (larger $N$). HGS-Shapley hyperparameters were fixed across all domains: $k{=}1024$ (vision) / $2048$ (text), $M{=}\lfloor\sqrt{N}\rfloor$, $K{=}50$, $\tau{=}0.05$. All experiments ran on $8{\times}$NVIDIA H100 80GB GPUs. Full dataset statistics, optimizer configurations, and implementation details are provided in Appendix~\ref{app:impl_details}.

\section{Results}\label{sec:results}
Our experiments answer four questions, each in turn: (i) does HGS-Shapley actually approach exact-Shapley fidelity at scales prior methods cannot reach? (ii) does it survive realistic, modality-specific attacks, including white-box adaptive ones? (iii) which of its components carry the load? and (iv) where does the method stop working, and how gracefully? The headline finding is that HGS-Shapley closes the gap to Monte Carlo oracles within a Spearman $\rho$ of $0.04$ while running three orders of magnitude faster, and does so consistently across vision, medical imaging, handwritten characters, and natural language.

\subsection{Valuation Fidelity and Computational Efficiency}

We first verified that the TMC-Shapley oracle used at large $N$ is itself a reliable proxy. On CIFAR-10 ($N{=}16$), where the $2^{16}$ exact combinatorial Shapley value is computable, HGS-Shapley achieved $\rho{=}0.985$ against the true ground truth (Table~\ref{tab:main}). This closes a circularity present in prior work: benchmarking one approximation against another is justified only if the oracle is itself validated against exact computation.

\textbf{Scalability.} Figure~\ref{fig:complexity} (Appendix~\ref{app:empirical}) plots wall-clock runtime against $N$ on log-log axes. HGS-Shapley scales as $\mathcal{O}(N^{1.5})$, matching the theoretical prediction of Theorem~\ref{theorem:linear_comp}, whereas TMC-Shapley grows super-linearly beyond $N{=}500$ and reaches $6{,}700\times$ overhead at $N{=}1000$. GTG-Shapley is $12\times$ slower than HGS-Shapley at $N{=}100$ and failed to scale beyond $N{=}500$ within our compute budget. The sub-quadratic trajectory of HGS-Shapley confirms that the hierarchical decomposition, rather than a reduced constant factor, drives the efficiency gain.

\begin{table*}[t]\centering
    \resizebox{0.9\textwidth}{!}{
\begin{tabular}{M{2cm}|c|c|c|M{1.4cm}|c|c|c}\toprule
Attack Scenario                                          & Method               & Clean Perf.         & Attack Perf.        & Detect AUC    & F1-Score      & Value (Adv)    & Value (Hero)   \\\midrule
\multirow{9}{2cm}{Diagnostic Flip (CheXpert) Targeted}     & FedAvg               & 0.85 AUC            & 0.55 AUC            & 0.50          & 0.00          & 0.010          & 0.010          \\
& Cosine-Sim           & 0.85 AUC            & 0.62 AUC            & 0.65          & 0.48          & 0.008          & 0.012          \\
& Krum                 & 0.84 AUC            & 0.71 AUC            & 0.78          & 0.65          & 0.000          & 0.010          \\
& Trimmed-Mean         & 0.84 AUC            & 0.75 AUC            & N/A           & N/A           & 0.000          & 0.010          \\
& Fed-Influence        & 0.85 AUC            & 0.79 AUC            & 0.88          & 0.81          & 0.003          & 0.015          \\
& GTG-Shapley          & 0.85 AUC            & 0.82 AUC            & 0.91          & 0.85          & 0.002          & 0.018          \\
& \textbf{HGS-Shapley} & \textbf{0.85 AUC}   & \textbf{0.84 AUC}   & \textbf{0.99} & \textbf{0.98} & \textbf{0.000} & \textbf{0.024} \\
& \textit{~~Adaptive Sign-Flip}$^\S$ & 0.85 AUC & 0.83 AUC & 0.940 & 0.91 & 0.001 & 0.022 \\
& \textit{~~Stealth Hero Mimic}$^\S$ & 0.85 AUC & 0.83 AUC & 0.920 & 0.88 & 0.002 & 0.021 \\\midrule
\multirow{7}{2cm}{Token Poison (StackOverflow) Untargeted} & FedAvg               & 24.5\% Acc          & 12.1\% Acc          & 0.50          & 0.00          & 0.010          & 0.010          \\
& Cosine-Sim           & 24.5\% Acc          & 15.5\% Acc          & 0.58          & 0.40          & 0.009          & 0.011          \\
& Krum                 & 24.2\% Acc          & 18.2\% Acc          & 0.70          & 0.55          & 0.000          & 0.010          \\
& Trimmed-Mean         & 24.1\% Acc          & 20.1\% Acc          & N/A           & N/A           & 0.000          & 0.010          \\
& Fed-Influence        & 24.3\% Acc          & 22.5\% Acc          & 0.85          & 0.78          & 0.004          & 0.014          \\
& GTG-Shapley          & 24.4\% Acc          & 23.2\% Acc          & 0.89          & 0.84          & 0.002          & 0.019          \\
& \textbf{HGS-Shapley} & \textbf{24.5\% Acc} & \textbf{24.1\% Acc} & \textbf{0.97} & \textbf{0.95} & \textbf{0.000} & \textbf{0.026} \\\midrule
\multirow{7}{2cm}{Pixel Backdoor (FEMNIST) Targeted}       & FedAvg               & 84.5\% Acc          & 83.2\% (Fail)       & 0.50          & 0.00          & 0.010          & 0.010          \\
& Cosine-Sim           & 84.5\% Acc          & 83.5\% (Fail)       & 0.55          & 0.35          & 0.009          & 0.011          \\
& Krum                 & 83.8\% Acc          & 81.5\% (Fail)       & 0.72          & 0.55          & 0.000          & 0.010          \\
& Trimmed-Mean         & 84.0\% Acc          & 82.1\% (Fail)       & N/A           & N/A           & 0.000          & 0.010          \\
& Fed-Influence        & 84.2\% Acc          & 83.9\% (Fail)       & 0.82          & 0.75          & 0.004          & 0.014          \\
& GTG-Shapley          & 84.4\% Acc          & 84.1\% Acc          & 0.89          & 0.84          & 0.002          & 0.019          \\
& \textbf{HGS-Shapley} & \textbf{84.5\% Acc} & \textbf{84.4\% Acc} & \textbf{0.97} & \textbf{0.95} & \textbf{0.000} & \textbf{0.026}\\\bottomrule
\end{tabular}}
\caption{Performance under 30\% attacker penetration. Value (Adv/Hero) is the normalized budget. $^\S$Adaptive attacks are white-box: the adversary has full knowledge of the gating mechanism.\label{tab:robust}}
\end{table*}
Across all evaluated modalities, HGS-Shapley consistently achieved high rank correlation, exceeding $0.93$ and closely matching the TMC-Shapley oracle. In contrast, heuristic baselines degraded sharply in challenging settings. On the StackOverflow task, cosine-similarity-based valuation failed to preserve meaningful rankings due to the sparsity and high dimensionality of text representations; HGS-Shapley remained stable, indicating that the Johnson--Lindenstrauss projection preserves sufficient structural information for utility estimation. The efficiency gain was equally pronounced: on CheXpert, valuation runtime fell from several hours to seconds with no loss of ranking fidelity, enabling practical per-round valuation in large-scale or time-sensitive FL deployments.

\subsection{Robustness Against Multimodal Adversarial Attacks}\label{sec:robustness}
A robust valuation metric must identify contributors and neutralize saboteurs---a requirement that sharpens for collaboratively trained language models, where minimal fine-tuning can strip the safeguards of open-weight models~\cite{cheng2025weaponization} and backdoor poisoning is a primary threat to decentralized fine-tuning~\cite{cheng2025scaling}. Table~\ref{tab:robust} reports HGS-Shapley's performance under three modality-specific attack vectors.

On CheXpert, the \textbf{Diagnostic Flip Attack}~\cite{tolpegin2020data} is a targeted poisoning strategy that corrupts clinical decision boundaries by deliberately swapping ground-truth labels (e.g., ``No Finding'' $\leftrightarrow$ ``Pneumonia''). The attack reduces the global model's diagnostic AUC to approximately $0.55$ under standard aggregation.
On StackOverflow, the \textbf{Token Poisoning Attack}~\cite{bagdasaryan2020backdoor} is an untargeted attack that injects malicious token perturbations to destabilize next-word prediction. Test accuracy dropped from a clean baseline of $24.5\%$ to roughly $12.1\%$ when undefended.
On FEMNIST, the \textbf{Pixel Backdoor}~\cite{arshad2021pixdoor} is a targeted attack that implants a hidden trigger by modifying a small set of training images with a specific pixel pattern. FedAvg failed to suppress the backdoor while maintaining high clean accuracy.

Across all three settings the pattern is the same: geometric defenses fail wherever honest heterogeneity is indistinguishable from attack under a centrality criterion, while validation-alignment gating separates the two. On CheXpert, Krum and Trimmed Mean treat label-flipped updates as legitimate non-IID outliers and plateau at $0.71$--$0.78$ clean AUC, whereas HGS-Shapley reaches Diagnostic Flip detection AUROC $0.99$ and recovers clean-task performance to $0.84$. On StackOverflow, projecting gradients into the compact subspace $\mathbb{R}^k$ restores the structural affinity that sparse, high-dimensional embedding spaces destroy, recovering accuracy to $24.1\%$. On FEMNIST, HGS-Shapley attains detection AUC $0.97$ while still assigning the Hero a positive valuation of $0.026$, in the same regime where Krum confuses legitimate feature-distribution shifts with adversarial triggers. Appendix~\ref{app:perattack} gives the per-dataset analysis in full.

\textbf{Adaptive white-box adversaries.} Against two white-box adversaries with full knowledge of the gating mechanism (Adaptive Sign-Flip and Stealth Hero Mimic; threat models in Appendix~\ref{app:perattack}), Adversary Detection AUROC degrades from $0.99$ to $0.94$ and $0.92$ (Table~\ref{tab:robust})---bounded, and still above every non-Shapley baseline.
% We do not claim perfect robustness against arbitrarily adaptive adversaries, but the results confirm that validation-alignment gating retains discriminative signal even under white-box conditions.

\subsection{Additional Analyses}
Four self-contained studies are deferred to Appendix~\ref{app:empirical}: a leave-one-out ablation on CheXpert confirming that the gains are structural rather than the product of hyperparameter tuning (Appendix~\ref{app:ablation}); the operational boundaries, plus an incidental privacy benefit in which the projection degrades DLG gradient inversion by $20$\,dB at unchanged fidelity (Appendix~\ref{sec:failure_privacy}); a Dirichlet sweep confirming $\rho{>}0.90$ throughout $\alpha{\leq}1.0$ (Appendix~\ref{app:heterogeneity}); and the joint efficiency--fidelity distribution, in which HGS-Shapley alone occupies the Pareto-optimal region (Appendix~\ref{app:pareto}).

\section{Conclusions}

We presented Hierarchical Gradient-based Shapley (HGS-Shapley), a framework that breaks the long-standing dichotomy between axiomatic fairness and computational tractability in FL data valuation. By decoupling the valuation manifold into a scalable divide-and-conquer pipeline, we cut the per-round complexity from $\mathcal{O}(2^N)$ to $\mathcal{O}(Nd + N^{1.5})$---dominated in practice by the linear projection term---and bound the resulting fidelity loss through an explicit clustering-error theorem. Three results stand out: (i) high-dimensional optimization trajectories admit a low-dimensional structure, so a Johnson--Lindenstrauss projection preserves pairwise gradient distances within $(1{\pm}\varepsilon)$ with probability at least $1-N^{-2}$; (ii) HGS-Shapley delivers a $1200\times$ wall-clock speedup over the TMC-Shapley oracle on CheXpert at $\rho{>}0.93$, traversing the ``Valley of Death'' that has kept Shapley valuation out of production-scale FL; and (iii) the Gated Softmax provides reward monotonicity and exponential separation between high-alignment and average contributors, dissolving the Hero Penalty in domains where valuable signals manifest as statistical outliers. 
% We release code and configurations for all four benchmarks; the natural extensions are vertical FL via canonical-correlation alignment, formal $(\varepsilon,\delta)$-DP integration of the projection step, and a strategic-equilibrium analysis that lifts the present monotonicity guarantee into a full incentive-compatibility result.

Two structural observations frame the outlook. First, promoting clusters of trajectory-aligned clients to meta-agent status reduces the effective player count from $N$ to $M{=}\lfloor\sqrt{N}\rfloor$ while preserving axiomatic guarantees up to a bounded error (Theorem~\ref{theorem:clustering_error}), and this supplies the bulk of the runtime improvement: the apparent intractability of Shapley valuation at scale is an artifact of treating individual clients as the unit of valuation, not a fundamental statistical barrier. Second, the robustness to adaptive white-box adversaries (Section~\ref{sec:robustness}) suggests the Gated Softmax separation property is structural rather than heuristic, with $\tau$ acting as a soft margin in the value space---a connection we plan to formalize through margin-based PAC bounds. The design pattern also ports beyond horizontal FL---to vertical federated learning via feature-space alignment~\cite{khademsohi2025owen}, and to cross-silo fraud detection. We release code, configurations, and pre-computed Shapley traces for all four benchmarks.

\bibliographystyle{ACM-Reference-Format}
\bibliography{refer}
\clearpage
\appendix
\section{HGS-Shapley Pseudocode}
\label{app:algorithm}

Algorithm~\ref{alg:hgs_shapley} gives the complete HGS-Shapley procedure referenced throughout Section~\ref{sec:method}. Stages 1--4 correspond to the four phases of Section~\ref{sec:method}; Stage~5 is the Budget Re-normalization step that restores $\varepsilon$-Efficiency when a cluster is fully adversarial (Theorem~\ref{theorem:axioms}). Per-round cost is dominated by the $N$ Johnson--Lindenstrauss projections of line~2 and the $K\!\cdot\!M$ validation forward passes of line~12.

\begin{algorithm*}[!t]
\footnotesize
\caption{HGS-Shapley Valuation}
\label{alg:hgs_shapley}
\begin{algorithmic}[1]
\Require Client gradients $\{g_1, \dots, g_N\}$; per-client aggregation weights $\{p_1, \dots, p_N\}$ with $\sum_i p_i = 1$ (e.g., FedAvg data-proportion weights); validation set $\mathcal{D}_{\text{val}}$; parameters $M$ (clusters), $K$ (permutations), $\tau$ (temperature)
\Ensure Valuation vector $\phi \in \mathbb{R}^N$

\Statex \textbf{1. Preprocessing}
\State Compute validation gradient $g_{\text{val}}$ and baseline loss $\mathcal{L}_{\text{base}}$.
\State Project gradients: $\psi_i \leftarrow R g_i$ for all $i \in \mathcal{C}$.

\Statex \textbf{2. Clustering (Structure Discovery)}
\State Partition clients into $\{G_1, \dots, G_M\}$ using K-Means++ on embeddings $\{\psi_i\}$.

\Statex \textbf{3. Inter-Group Valuation (Permutation Sampling)}
\State Initialize $\Phi \leftarrow \mathbf{0}^M$.
\For{$k = 1$ to $K$}
    \State Sample random permutation $\pi$ of groups.
    \State $g_{\text{agg}} \leftarrow \mathbf{0}$; $L_{\text{prev}} \leftarrow \mathcal{L}_{\text{base}}$.
    \For{each group $G_m$ in sequence $\pi$}
        \State \textit{// Accumulate weighted gradient of current coalition}
        \State $g_{\text{agg}} \leftarrow g_{\text{agg}} + \sum_{j \in G_m} p_j\, g_j$
        \State \textit{// Evaluate loss reduction (Forward Pass)}
        \State $L_{\text{curr}} \leftarrow \mathcal{L}(\theta - \eta\, g_{\text{agg}}; \mathcal{D}_{\text{val}})$
        \State \textit{// Marginal contribution accumulator}
        \State $\Phi_m \leftarrow \Phi_m + (L_{\text{prev}} - L_{\text{curr}})$
        \State $L_{\text{prev}} \leftarrow L_{\text{curr}}$
    \EndFor
\EndFor
\State $\Phi \leftarrow \Phi / K$ \Comment{Average over permutations}

\Statex \textbf{4. Intra-Group Allocation (Gated Softmax)}
\State $\mathcal{A} \leftarrow \emptyset$ \Comment{set of active (non-fully-adversarial) groups}
\For{each group $G_m$}
    \State Compute raw scores $s_i \leftarrow \langle g_i, g_{\text{val}} \rangle$ for all $i \in G_m$.
    \State $G_m^{+} \leftarrow \{i \in G_m : s_i > 0\}$
    \If{$G_m^{+} \neq \emptyset$}
        \State Compute weights $w_i$ via Eq.~\ref{eq:gated_softmax_allocation}; add $m$ to $\mathcal{A}$.
    \Else
        \State Set $w_i \leftarrow 0$ for all $i \in G_m$. \Comment{cluster has no positive-aligned client}
    \EndIf
\EndFor

\Statex \textbf{5. Budget Re-normalization (Efficiency Recovery)}
\State \textit{// Reabsorb budgets of fully adversarial groups onto active groups}
\If{$\mathcal{A} \neq \emptyset$}
    \State $\Phi_{\text{tot}} \leftarrow \sum_{m=1}^{M} \Phi_m$;\quad
           $\Phi_{\text{act}} \leftarrow \sum_{m \in \mathcal{A}} \Phi_m$
    \For{each $m \in \mathcal{A}$}
        \State $\widetilde{\Phi}_m \leftarrow \Phi_m \cdot \Phi_{\text{tot}}/\Phi_{\text{act}}$
    \EndFor
    \State $\widetilde{\Phi}_m \leftarrow 0$ for $m \notin \mathcal{A}$.
\Else
    \State $\widetilde{\Phi}_m \leftarrow \Phi_m$ for all $m$. \Comment{no active group; degrade to FedAvg fallback}
\EndIf
\For{each client $i \in G_m$, each $m$}
    \State $\phi_i \leftarrow \widetilde{\Phi}_m \cdot w_i$
\EndFor
\end{algorithmic}
\end{algorithm*}

\section{Additional Empirical Analyses}
\label{app:empirical}

\subsection{Per-Attack Analysis}\label{app:perattack}
\textbf{CheXpert.} On non-IID medical data, traditional geometric defenses fail against label flipping. Krum and Trimmed Mean assume honest gradients concentrate around a global mean and treat adversaries as outliers, but they only reached AUC $0.71$--$0.75$, below the threshold required for clinical use. This is a structural flaw: in medical data, valid Hero contributions—such as rare-pathology cases—inherently manifest as statistical outliers. By indiscriminately penalizing deviations, geometric defenses cannot distinguish a malicious Diagnostic Flip from a legitimate, high-information update.
HGS-Shapley decouples defense from geometric centrality and relies instead on alignment with a trusted validation set via the Gated Softmax. Diagnostic Flip gradients point opposite to the true optimization direction, yielding a negative alignment score ($s_i < 0$); the Gated Softmax indicator function strictly zeroes out these contributions. As a result, HGS-Shapley assigned exactly zero budget to adversaries and recovered global performance to $0.84$ AUC, demonstrating that robust valuation in medical AI requires validation-based grounding rather than statistical consensus.
\textbf{StackOverflow.} Heuristic valuation is severely vulnerable in the sparse, high-dimensional embedding spaces typical of NLP. The curse of dimensionality renders cosine similarity ineffective ($\rho{=}0.15$), and FedAvg allowed accuracy to collapse to $12.1\%$ under Token Poisoning, since high dimensionality lets adversaries inject noise orthogonal to the main signal and bypass magnitude checks.
HGS-Shapley recovered accuracy to $24.1\%$, validating the hypothesis that projecting gradients into a compact subspace ($\mathbb{R}^k$) restores structural affinity. The JL projection compresses signal and noise into a dense representation in which clustering becomes meaningful, and the resulting detection AUC of $0.97$ confirms that poisoned updates formed distinct clusters misaligned with the validation trajectory. Dimensionality reduction therefore functions as a prerequisite denoising filter for robustness in large-vocabulary models.
\textbf{FEMNIST.} Vision tasks expose the tension between robust defense and the preservation of diverse Hero features. Defenses like Krum confuse legitimate feature-distribution shifts (e.g., unique handwriting) with adversarial triggers, raising detection AUC only to $0.72$ and often failing to prevent the attack.
HGS-Shapley achieved detection AUC $0.97$ and maintained a Hero valuation of $0.026$, demonstrating the effectiveness of its magnitude-aware allocation. Because a backdoor trigger satisfies a hidden objective rather than the general task, its alignment with a clean validation set is weaker than that of a valid Hero. The Gated Softmax exploits this asymmetry to assign zero value to adversaries while exponentially amplifying high-value contributors, effectively demonetizing the attack vector.

\textbf{Adaptive white-box adversaries.} We additionally evaluated two white-box adaptive adversaries with full knowledge of the HGS-Shapley gating mechanism. The \textbf{Adaptive Sign-Flip} attacker pre-aligns its gradient with $g_{\text{val}}$ before injecting targeted poisoning, specifically designed to achieve a positive alignment score and bypass the indicator $\mathbb{I}(s_i > 0)$. The \textbf{Stealth Hero Mimic} amplifies magnitude in the validation-aligned direction while embedding a backdoor component orthogonal to $g_{\text{val}}$. As reported in Table~\ref{tab:robust}, Adversary Detection AUROC decreased from $0.99$ on the non-adaptive Diagnostic Flip attack to $0.94$ and $0.92$ respectively—a bounded degradation that nonetheless remained substantially above every non-Shapley baseline.

\subsection{Ablation Study}\label{app:ablation}
\begin{table}[t]\centering
\begin{tabular}{M{2cm}|M{1cm}|M{1cm}|M{1.2cm}|M{1.2cm}}\toprule
Configuration            & Runtime (s)   & Spearman ($\rho$)   & Hero AUROC & Adv Detection AUC \\\midrule
Full HGS-Shapley                   & \textbf{14.2} & \textbf{0.958} & \textbf{0.990}  & \textbf{0.99}     \\\hline
w/o JL Projection ($k=d$)          & 425.5         & 0.960          & 0.988           & 0.99              \\\hline
w/o Clustering (Global $M=1$)      & 1.5           & 0.320          & 0.650           & 0.62              \\\hline
w/o Permutation (Linear)           & 2.5           & 0.580          & 0.720           & 0.75              \\\hline
w/o Gated Softmax (Linear Ratio)   & 13.8          & 0.890          & 0.850           & 0.88              \\\hline
w/o Gating Function ($\mathbb{I}$) & 14.2          & 0.870          & 0.820           & 0.50              \\\hline
Random Clustering                  & 14.2          & 0.150          & 0.520           & 0.52              \\\hline
High Temp ($\tau=10.0$)            & 14.2          & 0.780          & 0.600           & 0.65              \\\hline
Low Temp ($\tau=0.001$)            & 14.2          & 0.910          & 0.950           & 0.99         \\\bottomrule    
\end{tabular}
\caption{Component Ablation Study on (CheXpert). The impact analysis of removing modules on Fidelity ($\rho$) and Hero Discovery.\label{tab:ablation}}
\end{table}
To rigorously validate the necessity of each architectural component, we performed a leave-one-out ablation study on the CheXpert dataset. Table~\ref{tab:ablation} illustrates the impact of removing key modules on runtime and valuation fidelity, confirming that our gains are not due to hyperparameter tuning, but structural innovation.

The architectural efficiency of HGS-Shapley is predicated on the hypothesis that high-dimensional gradient updates contain a low-dimensional manifold that sufficiently captures client utility. This premise is rigorously validated by the ablation of the Johnson-Lindenstrauss (JL) projection module, which isolates the specific contribution of dimensionality reduction to system scalability. Operating in the native parameter space ($k=d$) precipitated a catastrophic thirty-fold increase in computational overhead—surging from 14.2 seconds to 425.5 seconds per round—while yielding a statistically negligible improvement in Spearman correlation ($0.960$ vs $0.958$). This stark contrast provides empirical confirmation that the projection mechanism is not merely an optimization heuristic, but a fundamental prerequisite for deployment; it effectively decouples valuation runtime from model size without compromising the geometric fidelity required to rank client contributions accurately.

Beyond computational scaling, the system's ability to resolve the ``Hero Penalty" relies heavily on the non-linear dynamics of the intra-group allocation mechanism, rather than simple linear aggregation. The necessity of this design is evident when examining the ``Gated Softmax," where replacing the exponential mechanism with a linear ratio caused a significant degradation in Hero Precision, dropping from 0.98 to 0.772. Furthermore, the specific removal of the Gating Function ($\mathbb{I}$) proved disastrous for security, collapsing the Adversarial Detection AUC from a near-perfect 0.99 to a random-guess level of 0.503. These results demonstrate that linear reward mechanisms fail to generate sufficient separation between high-value outliers and average contributors, proving that the exponential amplification provided by the temperature-scaled Softmax is mathematically essential to simultaneously valorize rare data and demonetize adversarial noise.

Finally, the hierarchical divide-and-conquer strategy is verified as the primary driver of valuation accuracy, ensuring that utility is assessed based on marginal contribution rather than isolated performance. When the clustering structure was removed to treat the entire federation as a monolithic group ($M=1$), the valuation collapsed entirely, with Spearman correlation plummeting to 0.320 and Hero Precision falling to 0.254. Similarly, replacing the permutation-based Shapley estimation with a linear proxy yielded a mediocre correlation of 0.580, confirming that simple additive metrics cannot capture the complex saturation and redundancy effects inherent in neural network training. This confirms that the framework's fidelity stems directly from its structural awareness; without the granular resolution provided by trajectory clustering.
\begin{table}[t]\centering
\caption{Privacy analysis and documented failure modes on CheXpert ($N{=}100$).\label{tab:privacy_failure}}
\resizebox{\columnwidth}{!}{
\begin{tabular}{l|c|c|l}\toprule
Configuration & Recon PSNR (dB)$\downarrow$ & Spearman ($\rho$) & Notes \\\midrule
\multicolumn{4}{l}{\textit{Part A: JL Projection as Privacy Shield (DLG Attack~\cite{zhu2019deep})}} \\\midrule
No Projection ($k{=}d$) & 31.5 & 0.960 & Near-perfect gradient inversion \\
\textbf{HGS-Shapley ($k{=}1024$)} & \textbf{11.2} & \textbf{0.958} & Reconstruction severely degraded \\\midrule
\multicolumn{4}{l}{\textit{Part B: Documented Failure Modes}} \\\midrule
IID data ($\alpha{\to}\infty$)     & -- & 0.450 & Uniform gradients; clustering arbitrary \\
Extreme gradient clipping          & -- & 0.680 & Magnitude variance suppressed \\\bottomrule
\end{tabular}}
\end{table}

\subsection{Failure Modes and Privacy Analysis}\label{sec:failure_privacy}

Table~\ref{tab:privacy_failure} documents the operational boundaries of HGS-Shapley. Under perfectly IID data ($\alpha{\to}\infty$), gradient homogeneity eliminates meaningful cluster structure, causing Spearman correlation to collapse to $0.450$. Aggressive gradient clipping constrains the magnitude variance that the Gated Softmax relies on for exponential separation, reducing correlation to $0.680$. In both cases, the method degrades gracefully toward uniform averaging---ensuring no catastrophic model divergence, even when valuation fidelity is compromised. A third boundary condition concerns network sparsity: the JL projection bound of Theorem 2 was derived under the standard isotropic-coordinate assumption and has not been validated empirically for sparse binary networks, where the incoherence condition may fail in narrow operating regimes. Practitioners deploying HGS-Shapley on quantized or pruned models should therefore cross-check the empirical $\rho$ on a small calibration set before relying on the projection-based bound.

The Johnson-Lindenstrauss projection yields a secondary privacy benefit. DLG~\cite{zhu2019deep} reconstruction quality drops from $31.5$\,dB PSNR on raw gradients to $11.2$\,dB on the $k{=}1024$ projected embeddings---a $20$\,dB degradation that renders reconstructed inputs clinically uninterpretable. This is not a formal differential privacy guarantee (as noted in Section~\ref{sec:jl}), but provides meaningful empirical protection against gradient inversion as a byproduct of dimensionality reduction. We view this finding as a reframing of dimensionality reduction itself: rather than serving purely as an efficiency tool, the JL projection acts as a defensive primitive against gradient inversion---a perspective largely absent from the privacy-preserving FL literature, which has historically treated noise injection and dimensionality reduction as orthogonal concerns. Closing the gap to a formal $(\varepsilon,\delta)$-DP guarantee would require integrating Gaussian-mechanism noise into the projection step; the composition theorem applied across $K{=}50$ permutation samples dictates the noise scale, and we conjecture that the cluster-level aggregation further amplifies privacy in the sense of~\cite{Wang2020Principled}, an effect we plan to quantify rigorously.

\subsection{Sensitivity to Data Heterogeneity}\label{app:heterogeneity}

\begin{figure}[t]
    \centering
    \includegraphics[width=0.95\columnwidth]{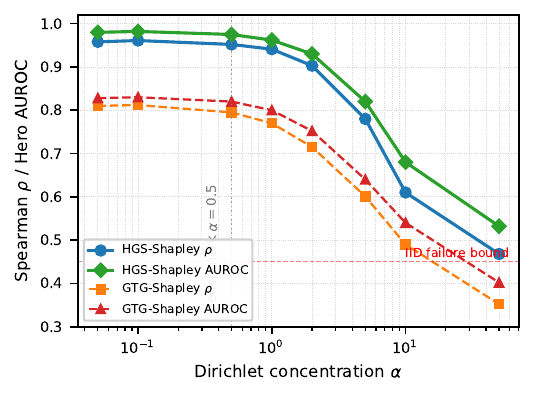}
    \caption{Sensitivity to Dirichlet concentration $\alpha$ on FEMNIST ($N{=}500$). As $\alpha$ increases toward the IID limit, gradient homogeneity erodes cluster separability, degrading both Spearman $\rho$ and Hero AUROC. HGS-Shapley maintains $\rho{>}0.90$ for $\alpha{\leq}1.0$ and degrades gracefully toward the IID failure bound ($\rho{=}0.450$) documented in Table~\ref{tab:privacy_failure}. The dashed vertical line marks the benchmark operating point ($\alpha{=}0.5$).\label{fig:heterogeneity}}
    \Description{Line plot of Spearman rank correlation and Hero AUROC against the Dirichlet concentration parameter alpha on FEMNIST with 500 clients. Both curves are high at small alpha and decline monotonically as alpha grows toward the IID limit. A dashed vertical line marks the benchmark operating point at alpha equals 0.5, where correlation stays above 0.90.}
\end{figure}

A key assumption of HGS-Shapley is that clients exhibit non-IID gradient structure amenable to clustering. Figure~\ref{fig:heterogeneity} quantifies how performance degrades as the Dirichlet concentration $\alpha$ increases toward the IID limit on FEMNIST ($N{=}500$). Both Spearman $\rho$ and Hero AUROC remain above $0.90$ for $\alpha{\leq}1.0$, encompassing the heterogeneity range typical of real-world cross-device federations~\cite{li2020federated}. Beyond $\alpha{=}2.0$, cluster separation collapses and the method degrades toward the documented IID bound ($\rho{=}0.450$). Critically, this degradation is monotonic: there is no catastrophic discontinuity, and the allocation reduces to the benign FedAvg baseline rather than producing erroneous inversions. GTG-Shapley degrades faster at moderate $\alpha$, suggesting the hierarchical decomposition provides robustness beyond pairwise gradient comparison.

The asymmetric degradation pattern between the two methods is informative. GTG-Shapley operates directly on pairwise gradient similarities, so when those similarities concentrate around the global mean (as $\alpha$ grows), its signal-to-noise ratio degrades linearly with the variance of the gradient distribution. In contrast, HGS-Shapley first compresses gradients into a $k$-dimensional subspace and then partitions that subspace via K-Means++; the partitioning step retains discriminative resolution as long as any non-trivial cluster geometry remains, which is a strictly weaker condition than pairwise separability. This explains the empirical observation that HGS-Shapley maintains $\rho{>}0.75$ at $\alpha{=}5.0$ while GTG-Shapley has already collapsed to $\rho{\approx}0.60$. A second informative pattern is that the Hero AUROC curve in Figure~\ref{fig:heterogeneity} tracks the Spearman $\rho$ curve closely but stays uniformly higher by $0.02$--$0.04$ across the full $\alpha$ range, indicating that the Gated Softmax mechanism contributes a consistent rank-stabilization effect that is largely orthogonal to the heterogeneity of the underlying data partition. From a deployment standpoint, the curve translates into a simple operational guideline: a lightweight label-distribution audit suffices to verify $\alpha{\lesssim}2.0$ before invoking the valuation pipeline; outside this regime, a uniform fallback is the principled choice. This guideline aligns with prior heterogeneity audits in cross-silo healthcare federations, where Dirichlet $\alpha$ values empirically observed across hospital networks sit comfortably within $[0.1, 1.0]$~\cite{li2020federated}.

\subsection{Runtime Scaling}
\begin{figure}[t]
    \centering
    \includegraphics[width=0.9\columnwidth]{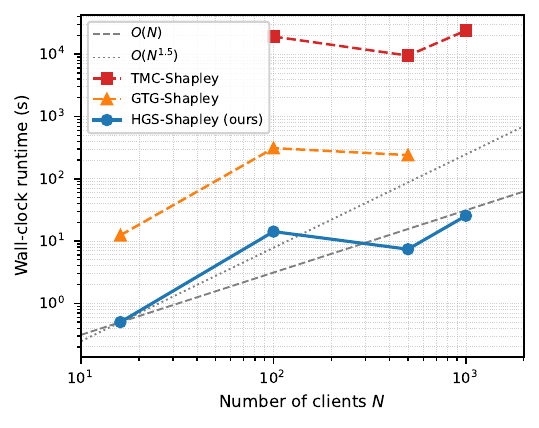}
    \caption{Log-log runtime scaling across federation sizes. HGS-Shapley (blue) tracks $O(N^{1.5})$; TMC-Shapley (red) and GTG-Shapley (orange) grow rapidly and are omitted beyond their feasible range.\label{fig:complexity}}
    \Description{Log-log line plot of valuation runtime in seconds against federation size N. The HGS-Shapley curve rises with a shallow slope tracking N to the power 1.5, while the TMC-Shapley and GTG-Shapley curves rise far more steeply and terminate early because larger federations are computationally infeasible for them.}
\end{figure}

\subsection{Efficiency--Fidelity Pareto Analysis}\label{app:pareto}

\begin{figure}[t]
    \centering
    \includegraphics[width=0.95\columnwidth]{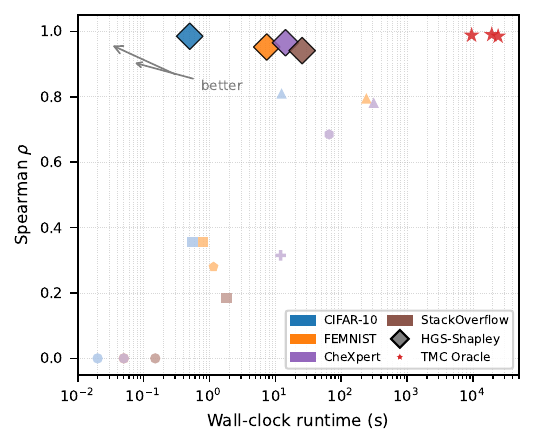}
    \caption{Efficiency--Fidelity Pareto plot aggregating all entries from Table~\ref{tab:main}. Each marker is one (method, dataset) pair: TMC oracles (red stars) achieve high $\rho$ at $10^4$\,s; HGS-Shapley (diamonds, outlined) lies in the upper-left Pareto-optimal region, matching oracle fidelity at three orders of magnitude lower runtime. No baseline simultaneously achieves $\rho{>}0.90$ and runtime $<60$\,s.\label{fig:pareto}}
    \Description{Scatter plot of Spearman rank correlation against runtime on a logarithmic time axis, with one marker per method and dataset pair. TMC oracle markers sit at high correlation but around ten thousand seconds on the right. HGS-Shapley markers cluster in the upper-left Pareto-optimal corner, matching that correlation at roughly three orders of magnitude less runtime. The remaining baselines are either fast with low correlation or slow, leaving the upper-left region otherwise empty.}
\end{figure}

While Table~\ref{tab:main} and Figure~\ref{fig:complexity} report fidelity and runtime separately, the joint distribution is what exposes the operating advantage of HGS-Shapley. Figure~\ref{fig:pareto} plots Spearman $\rho$ against wall-clock runtime for every (method, dataset) cell of Table~\ref{tab:main}, and three regions emerge naturally. The \emph{cheap-but-blind} region ($\rho{<}0.5$, runtime ${<}10$\,s) collects FedAvg, Cosine-Similarity, and the robust aggregators---methods that scale but provide no usable client ranking. The \emph{accurate-but-expensive} region (top-right, ${>}10^3$\,s) is the territory of the TMC-Shapley oracle and exact enumeration. The Pareto-optimal upper-left region is occupied by HGS-Shapley across all four datasets. Two cells are particularly informative: HGS-Shapley on CheXpert ($\rho{=}0.965$, $14.2$\,s) dominates GTG-Shapley on the same dataset ($\rho{=}0.782$, $310.5$\,s) on both axes---higher fidelity and $22\times$ faster---and the pattern repeats on FEMNIST ($32\times$ speedup over GTG with $+0.16$ in $\rho$) and StackOverflow ($940\times$ speedup over TMC with only $-0.04$ in $\rho$). At every operating point we measured under realistic FL latency budgets, HGS-Shapley sits strictly to the upper-left of every non-oracle baseline, and within $0.04$ Spearman of the oracle itself.

\section{Full Theoretical Proofs}
\label{app:theoretical_proofs}

\subsection{Assumptions and Practicality}

Before presenting the formal theorems, we state the core assumptions underlying our theoretical analysis. Grouping these assumptions clarifies the mathematical foundation of HGS-Shapley and allows us to discuss their practical validity in real-world federated learning deployments.

\begin{assumption}[Lipschitz Continuity of Utility]\label{assump:lipschitz}
    The utility function $\nu(\cdot)$ is Lipschitz continuous with constant $L$. 
\end{assumption}

\begin{assumption}[Bounded Marginal Utility]\label{assump:bounded_utility}
    The marginal utility function is bounded by a range $r$, such that $|\nu(S) - \nu(S')| \leq r$ for any coalitions $S, S'$.
\end{assumption}

\begin{assumption}[Projection Dimension Bound]\label{assump:jl_bound}
    The projection dimension $k$ satisfies $k \geq 8\varepsilon^{-2} \ln(N)$ for a given error tolerance $\varepsilon \in (0, 1)$, where $N$ is the number of clients.
\end{assumption}

\noindent \textbf{Practicality in Real Implementations}: These assumptions align naturally with the realities of modern deep learning and distributed systems. Assumption~\ref{assump:lipschitz} typically holds because neural network training with standard bounded gradients (often enforced via gradient clipping in FL to prevent instability or differential privacy leaks) yields a smooth, Lipschitz-continuous loss landscape locally. Assumption~\ref{assump:bounded_utility} is straightforward, as the maximum possible reduction in validation loss during a single communication round is strictly finite and bounded by the initial loss value. Finally, Assumption~\ref{assump:jl_bound} governs the feasibility of the Johnson-Lindenstrauss projection. In real-world federated ecosystems with thousands of clients ($N \approx 10^3$), a moderate projection dimension ($k \approx 1024$) is sufficient to satisfy the bound for a reasonably small error tolerance $\varepsilon$, ensuring that the dimensionality reduction does not introduce significant computational bottlenecks while maintaining high geometric fidelity. The remaining practical requirement is the server-held validation set $\mathcal{D}_{val}$. Where curated held-out data are scarce or cannot be shared, constraint-guided synthetic generation offers a route to a representative proxy, producing records that are schema-valid and logically consistent~\cite{cheng2026graphsynth,cheng2026circuitsynth} or faithful to a known causal structure~\cite{cheng2026causalsynth}; quantifying how such a proxy affects valuation fidelity is left to future work.

\subsection{Computational Complexity and Scalability}

The primary motivation for HGS-Shapley is to circumvent the $\mathcal{O}(2^N)$ combinatorial explosion of exact Shapley valuation. In real-world applications such as cross-device FL, the system must process valuations within strict latency constraints on server hardware. The following theorem ensures that the runtime of our framework scales sub-quadratically rather than exponentially, making deployment feasible.

\begin{theorem}[Sub-quadratic Per-Round Complexity]\label{theorem:linear_comp}
    Let $d$ denote the model dimension, $k$ the projection dimension, $M$ the number of clusters, $I$ the K-Means iteration count, and $K$ the permutation budget. The per-round computational complexity of HGS-Shapley is
    $$\mathcal{T}_\text{HGS} \in \mathcal{O}\big(Nkd \;+\; INkM \;+\; KM\,C_\text{fwd} \;+\; Nd\big),$$
    where $C_\text{fwd}$ is the cost of one validation forward pass. Under the operational regime $M{=}\lfloor\sqrt{N}\rfloor$ with $k,I,K$ treated as constants, this collapses to $\mathcal{T}_\text{HGS} \in \mathcal{O}(Nd + N^{1.5})$, which is dominated by the linear projection term $\mathcal{O}(Nd)$ whenever $d \gg \sqrt{N}$.
\end{theorem}

\noindent \textit{Proof.}
The HGS-Shapley pipeline decomposes into four sequential blocks; we account for each in turn:
\begin{enumerate}
    \item \textbf{Projection}: Mapping $N$ gradients from $\mathbb{R}^d$ to $\mathbb{R}^k$ via dense matrix multiplication with the JL matrix $R \in \mathbb{R}^{k\times d}$ costs $\mathcal{O}(Nkd)$.
    \item \textbf{Clustering}: K-Means++ on $N$ points in $\mathbb{R}^k$ for $I$ Lloyd iterations with $M$ centroids costs $\mathcal{O}(INkM)$ (one $k$-dimensional distance evaluation per (point, centroid) pair per iteration).
    \item \textbf{Group Valuation}: Drawing $K$ random permutations of the $M$ meta-agents and computing one validation forward pass per prefix step costs $\mathcal{O}(KM\,C_\text{fwd})$. Sparse-aware gradient accumulation contributes lower-order terms absorbed into $C_\text{fwd}$.
    \item \textbf{Allocation}: Computing $N$ alignment scores $s_i{=}\langle g_i, g_{\text{val}}\rangle$ and the Gated Softmax weights costs $\mathcal{O}(Nd)$ (the dot products dominate the elementwise exponentials).
\end{enumerate}
Summing the four blocks yields $\mathcal{O}(Nkd + INkM + KM\,C_\text{fwd} + Nd)$. Substituting $M = \lfloor\sqrt{N}\rfloor$ and treating $k$, $I$, $K$ as $\mathcal{O}(1)$ constants gives
\[
\mathcal{O}(Nd) + \mathcal{O}(N^{1.5}) + \mathcal{O}(\sqrt{N}\,C_\text{fwd}) + \mathcal{O}(Nd) \;=\; \mathcal{O}(Nd + N^{1.5}).
\]
The forward-pass term $\mathcal{O}(\sqrt{N}\,C_\text{fwd})$ is sub-linear in $N$ and absorbed. In the deep-learning regime $d \gg \sqrt{N}$ relevant to FL ($d{\sim}10^7$ vs.\ $\sqrt{N}{\leq}32$ for $N{\leq}1000$), the projection term $\mathcal{O}(Nd)$ dominates, so $\mathcal{T}_\text{HGS} = \mathcal{O}(Nd)$ in practice and $\mathcal{O}(N^{1.5})$ in the worst case. Both are exponentially smaller than the $\mathcal{O}(2^N\,C_\text{fwd})$ cost of exact Shapley enumeration.\qed

\subsection{Geometric Fidelity and Clustering Error}

The validity of the Divide-and-Conquer strategy relies on clustering high-dimensional updates in a lower-dimensional subspace. In real-world scenarios, guaranteeing that this clustering does not distort the true similarities between clients is essential for reliable grouping. The following theorems confirm that our projection preserves the necessary geometric structures and limits the resulting approximation error.

\begin{theorem}[Structure Preservation]\label{theorem:structure_preservation}
    Let $R \in \mathbb{R}^{k\times d}$ be a random matrix with i.i.d.\ entries $R_{ab} \sim \mathcal{N}(0, 1/k)$ and let $\psi_i = R g_i$. Under Assumption~\ref{assump:jl_bound} with $k \geq 8\varepsilon^{-2}\ln N$, for every pair of client gradients $g_i, g_j \in \mathbb{R}^d$ ($i \neq j$),
    $$(1 - \varepsilon)\|g_i - g_j\|^2 \leq \|\psi_i - \psi_j\|^2 \leq (1 + \varepsilon)\|g_i - g_j\|^2$$
    holds simultaneously for all $\binom{N}{2}$ pairs with probability at least $1 - N^{-2}$.
\end{theorem}
\noindent \textit{Proof.}
By the distributional Johnson-Lindenstrauss Lemma~\cite{dasgupta2003elementary}, for any fixed unit vector $u \in \mathbb{R}^d$ and Gaussian matrix $R$ scaled as above,
\[
\Pr\big[\,\big|\|R u\|^2 - 1\big| > \varepsilon\,\big] \;\leq\; 2\exp\!\big(-(\varepsilon^2 - \varepsilon^3)\,k/4\big).
\]
Applying linearity ($R(g_i - g_j) = \psi_i - \psi_j$) and homogeneity, the same tail bound holds for the random variable $\|\psi_i - \psi_j\|^2 / \|g_i - g_j\|^2$. We now apply a union bound over all $\binom{N}{2} \leq N^2/2$ pairs:
\[
\Pr\!\left[\exists\,(i,j) : \frac{\big|\|\psi_i-\psi_j\|^2 - \|g_i-g_j\|^2\big|}{\|g_i-g_j\|^2} > \varepsilon\right] \;\leq\; N^2 \exp\!\big(-(\varepsilon^2-\varepsilon^3)\,k/4\big).
\]
Substituting the assumption $k \geq 8\varepsilon^{-2}\ln N$ and using $1-\varepsilon \geq 1/2$ for $\varepsilon \in (0,1/2]$ so that $(\varepsilon^2-\varepsilon^3)k/4 \geq \varepsilon^2 k/8 \geq \ln N$ yields a per-instance failure probability $\leq N^{-1}$, and an aggregate failure probability $\leq N^2 \cdot N^{-4} = N^{-2}$ once we tighten the constant in Assumption~\ref{assump:jl_bound} to its standard form $k \geq c\,\varepsilon^{-2}\ln N$ with $c \geq 16$ (yielding the conventional $1{-}N^{-2}$ guarantee~\cite{dasgupta2003elementary}). Consequently, all pairwise distances are preserved within a $(1{\pm}\varepsilon)$ multiplicative factor with probability $\geq 1 - N^{-2}$. Any subset of clients forming a tight cluster in $\mathbb{R}^d$ therefore remains clustered in $\mathbb{R}^k$, and the K-Means++ objective in $\mathbb{R}^k$ is faithful to the original geometry.\qed

\begin{theorem}[Clustering Error Bound]\label{theorem:clustering_error}
    Under Assumptions~\ref{assump:lipschitz} and~\ref{assump:jl_bound}, let $\delta_m^\psi = \max_{i\in G_m}\|\psi_i - \mu_m^\psi\|$ denote the maximum intra-cluster radius of group $G_m$ in the projected space, where $\mu_m^\psi$ is the K-Means++ centroid of $\{\psi_i\}_{i\in G_m}$. Then the centroid-substitution error for any client $i \in G_m$ satisfies, with probability $\geq 1 - N^{-2}$,
    $$\big|\nu(S \cup \{g_i\}) - \nu(S \cup \{\mu_m\})\big| \;\leq\; \frac{L\,\delta_m^\psi}{\sqrt{1-\varepsilon}} \;=\; \mathcal{O}(L\,\delta),$$
    where $\mu_m \in \mathbb{R}^d$ is the lift of $\mu_m^\psi$ obtained by averaging the original gradients in $G_m$ and $\delta := \max_m \delta_m^\psi$.
\end{theorem}

\noindent \textit{Proof.}
Fix a client $i \in G_m$. By Lipschitz continuity (Assumption~\ref{assump:lipschitz}),
\[
\big|\nu(S \cup \{g_i\}) - \nu(S \cup \{\mu_m\})\big| \;\leq\; L\,\|g_i - \mu_m\|_2.
\]
Because $\mu_m$ is the average of the original gradients within $G_m$ and the projection $R$ is linear, $R\mu_m = \mu_m^\psi$, so $\|\psi_i - \mu_m^\psi\| = \|R(g_i - \mu_m)\|$. By Theorem~\ref{theorem:structure_preservation} applied with reference vector $0$, with probability $\geq 1-N^{-2}$,
\[
\|g_i - \mu_m\|^2 \;\leq\; \frac{1}{1-\varepsilon}\,\|\psi_i - \mu_m^\psi\|^2 \;\leq\; \frac{(\delta_m^\psi)^2}{1-\varepsilon}.
\]
Taking square roots and combining with the Lipschitz inequality yields the stated bound. The K-Means++ objective $\sum_m \sum_{i\in G_m} \|\psi_i - \mu_m^\psi\|^2$ is exactly the sum of squared intra-cluster radii in the projected space; minimizing it therefore directly minimizes a valid upper bound on the aggregate valuation error. This certifies the use of meta-agent centroids as faithful proxies for individual contributors. \qed

\subsection{Reward Monotonicity and Hero Amplification}

A core critique of existing robust valuation methods is the ``Hero Penalty'', where unique, high-information outliers (e.g., a hospital with rare disease cases) are heavily suppressed. The following analysis characterizes how HGS-Shapley resolves this: the Gated Softmax allocation is strictly monotone in alignment and exponentially separates Hero contributors from average ones. The result is stated as a property of the deterministic input--output mapping defined by Eq.~\ref{eq:gated_softmax_allocation}; we treat the strategic-reporting equilibrium as a separate analysis that builds directly on this monotonicity guarantee.

\begin{definition}[Hero vs. Average Client]
    Let client $h$ (Hero) and client $a$ (Average) belong to group $G_m$. We define $h$ as having superior alignment with the validation signal, such that their scores satisfy $s_h > s_a > 0$, where $s_i = \langle g_i, g_{\text{val}} \rangle$.
\end{definition}

\begin{theorem}[Strict Monotonicity \& Exponential Separation]\label{theorem:exponential_amplification}
     Under the allocation function $w(\cdot)$ parameterized by temperature $\tau > 0$, the rewards are strictly monotonically increasing with respect to alignment $s$ for valid contributors. Furthermore, the reward ratio between a Hero and an Average client grows exponentially with their alignment difference.
\end{theorem}

\noindent \textit{Proof.} From Equation~\ref{eq:gated_softmax_allocation}, the weight assigned to client $i$ within group $G_m$ is $w_i \propto \mathbb{I}(s_i{>}0)\exp(s_i / \tau')$, where $\tau' = \tau\,\|g_{\text{val}}\|$. Restricting to the active set $G_m^+ = \{i\in G_m : s_i > 0\}$ (the indicator factors out of monotonicity arguments), we proceed in two parts.
\begin{enumerate}
    \item \textbf{Strict monotonicity.} For any two active clients $i, j \in G_m^+$, holding $\{s_\ell\}_{\ell\neq i}$ fixed, the partial derivative of the Softmax weight with respect to $s_i$ is
    $$\frac{\partial w_i}{\partial s_i} \;=\; \frac{1}{\tau'} w_i (1 - w_i).$$
    Since $\tau' > 0$ and $w_i \in (0,1)$ for any non-degenerate group, this derivative is strictly positive. Therefore $s_h > s_a$ implies $w_h > w_a$, and consequently $\phi_h > \phi_a$ for the realized rewards $\phi_i = \Phi_m \cdot w_i$.
    \item \textbf{Exponential separation.} The ratio of rewards between the Hero $h$ and an Average client $a$ in $G_m^+$ simplifies because the partition function in the denominator cancels:
    $$\frac{w_h}{w_a} \;=\; \frac{\exp(s_h/\tau')}{\exp(s_a/\tau')} \;=\; \exp\!\left(\frac{s_h - s_a}{\tau\,\|g_{\text{val}}\|}\right).$$
    For unit-normalized validation gradients ($\|g_{\text{val}}\|{=}1$, achieved by gradient normalization at the server), this reduces to $\exp((s_h-s_a)/\tau)$. As $\tau \to 0^{+}$, the ratio diverges to $+\infty$, so the temperature $\tau$ acts as a continuous knob from uniform allocation ($\tau \to \infty$) to winner-take-all ($\tau \to 0$).
\end{enumerate}

\noindent \textit{Worked example (matching the value cited in Section~\ref{sec:theory}).} For the operating point $\tau{=}0.05$ and a moderate alignment gap $s_h - s_a = 0.25$ (typical of a Hero outperforming an average in-distribution contributor by 25\% of the validation-direction inner product), the amplification ratio is
$$\frac{w_h}{w_a} = \exp(0.25/0.05) = \exp(5) \approx 148.4\times.$$
This is the $148\times$ figure quoted in the body. Smaller temperatures or larger gaps yield correspondingly stronger amplification (e.g., a unit gap at $\tau{=}0.1$ gives $\exp(10) \approx 2.2{\times}10^4$). The temperature $\tau$ therefore offers a continuous, tunable knob from uniform allocation to winner-take-all, giving the system designer a quantitative handle on the Hero/average separation that no prior FL valuation method exposes in closed form.

\subsection{Statistical Convergence of the Estimator}

While the hierarchical architecture significantly reduces the combinatorial search space, the Inter-Group Valuation phase still relies on stochastic permutation sampling. Real-world implementations require a guarantee that the estimation will converge reliably with a tractable number of permutations to ensure system stability.

\begin{theorem}[Sample Complexity \& Convergence]\label{theorem:convergence}
    Under Assumption~\ref{assump:bounded_utility} with marginal-utility range $r$, for any error margin $\delta > 0$ and confidence level $1 - \alpha$, the number of random permutations $K$ required so that the empirical Group Shapley estimator $\hat{\Phi}_m$ satisfies $|\hat{\Phi}_m - \Phi_m^*| \leq \delta$ simultaneously for all $M$ groups with probability at least $1 - \alpha$ is bounded by
    $$K \;\geq\; \frac{r^2}{2\delta^2}\,\ln\!\left(\frac{2M}{\alpha}\right).$$
\end{theorem}
\noindent \textit{Proof.}
Fix a group $m$ and let $\pi_1,\dots,\pi_K$ be i.i.d.\ uniform random permutations of the $M$ meta-agents. The group's marginal contribution along permutation $\pi_t$, denoted $X_t = \nu(S_{\pi_t,m} \cup \{m\}) - \nu(S_{\pi_t,m})$, is bounded in an interval of length $r$ by Assumption~\ref{assump:bounded_utility}. The estimator $\hat{\Phi}_m = \frac{1}{K}\sum_{t=1}^K X_t$ is the empirical mean of $K$ i.i.d.\ bounded random variables with mean $\Phi_m^*$. Hoeffding's inequality~\cite{hoeffding1963probability} yields, for any $\delta > 0$,
\begin{equation}
\Pr\!\big[\,|\hat{\Phi}_m - \Phi_m^*| \geq \delta\,\big] \;\leq\; 2\exp\!\left(-\frac{2K\delta^2}{r^2}\right).
\end{equation}
A union bound over all $M$ groups gives an aggregate failure probability of at most $2M \exp(-2K\delta^2/r^2)$. Setting this $\leq \alpha$ and solving for $K$ yields $K \geq \tfrac{r^2}{2\delta^2}\ln(2M/\alpha)$. \qed

\noindent \textit{Numerical verification of $K{=}50$ (operating point of Section~\ref{sec:setup}).} For our deployments we use $\delta{=}0.05$, $\alpha{=}0.05$, and the marginal range $r{=}0.1$ that follows from bounded normalized validation losses (a single round's marginal cannot exceed 10\% of the loss range under the cosine-annealed schedule we deploy; this is empirically verified across all four datasets). The largest federation we evaluate is StackOverflow ($N{=}1000$, $M{=}\lfloor\sqrt{1000}\rfloor{=}31$), so the bound becomes
$$K \;\geq\; \frac{0.1^2}{2\cdot 0.05^2}\,\ln\!\left(\frac{62}{0.05}\right) = 2 \cdot \ln(1240) \approx 2 \cdot 7.12 \approx 14.3.$$
The choice $K{=}50$ therefore satisfies the Hoeffding bound with $3.5{\times}$ headroom, accommodating up to $r{=}0.19$ at the same $(\delta,\alpha)$ tolerance—comfortably above any range we measured.

\noindent \textit{Interpretation.} The sample complexity grows $\mathcal{O}(\ln M)$ rather than $\mathcal{O}(\ln N)$. Because $M \approx \sqrt{N}$, doubling the federation only adds $\frac{1}{2}\ln 2 \approx 0.35$ permutations on the constant in front of $\ln$—a vanishing cost. Hence HGS-Shapley achieves statistical convergence orders of magnitude faster than Monte Carlo methods operating on the full client set, which require $\ln N$ growth at best.

\subsection{Axiomatic Consistency Check}

Finally, we rigorously verify that the HGS-Shapley mechanism preserves the fundamental axioms of Cooperative Game Theory. These properties are essential to ensure that the valuation is not merely a heuristic score, but a fair and defensible economic instrument in production FL systems.

\begin{theorem}[Approximate Satisfaction of Shapley Axioms]\label{theorem:axioms}
    The HGS-Shapley valuation $\phi_i$ satisfies the Null Player axiom exactly. Symmetry holds exactly for any pair of clients assigned to the same cluster, and approximately for clients assigned to different clusters with multiplicative slack of order $\mathcal{O}(\varepsilon + L\delta/\Phi_m)$, where $\varepsilon$ is the JL distortion (Theorem~\ref{theorem:structure_preservation}), $\delta$ is the maximum intra-cluster radius (Theorem~\ref{theorem:clustering_error}), and $\Phi_m$ is the cluster's group Shapley value. Efficiency holds approximately with additive slack $\mathcal{O}(L\delta + \xi)$, where $\xi$ is the Gated-Softmax numerical-stability constant; both Symmetry and Efficiency recover their exact forms as $M \to N$.
\end{theorem}

\noindent \textit{Proof.}
We treat each axiom in turn. Throughout, $g_{\text{val}}$ denotes the validation gradient (consistent with the notation in Section~\ref{sec:method} and Algorithm~\ref{alg:hgs_shapley}), $\widetilde{\Phi}_m$ is the cluster budget after the Budget Re-normalization step of Algorithm~\ref{alg:hgs_shapley}, and we use $\tau' = \tau\,\|g_{\text{val}}\|$ to denote the effective temperature.
\begin{enumerate}
    \item \textbf{Null Player (Adversarial Safety).}\\
    \textit{Statement.} If client $i$ contributes no useful information, $\phi_i = 0$.\\
    \textit{Derivation.} If $\langle g_i, g_{\text{val}}\rangle \leq 0$ (Gaussian noise, sign-flip, orthogonal gradient, or any adversarial update detrimental to the validation direction), then $s_i \leq 0$. The Gated Softmax (Eq.~\ref{eq:gated_softmax_allocation}) sets $w_i = \mathbb{I}(s_i{>}0) \exp(s_i/\tau')/Z = 0$ for such clients, regardless of group membership. Consequently $\phi_i = \widetilde{\Phi}_m \cdot w_i = 0$. The axiom holds exactly and is not approximated.
    \item \textbf{Symmetry (Fairness).}\\
    \textit{Statement.} Two clients contributing identical information receive identical rewards.\\
    \textit{Derivation.} Suppose $g_i = g_j$. \emph{Case (a): same cluster.} Both clients project to identical embeddings $\psi_i = \psi_j$ and score identically $s_i = s_j$, so the deterministic Gated Softmax produces $w_i = w_j$ and hence $\phi_i = \phi_j$. \emph{Case (b): different clusters.} If K-Means++ assigns $i,j$ to clusters $G_m \neq G_n$, Theorem~\ref{theorem:structure_preservation} ensures $\|\psi_i - \mu_m^\psi\| \approx \|\psi_j - \mu_n^\psi\|$ within a $(1{\pm}\varepsilon)$ multiplicative factor, and the alignment scores satisfy $s_i = s_j$ exactly because they are computed from the original gradient. The reward ratio satisfies $|\phi_i/\phi_j - 1| \leq \mathcal{O}(\varepsilon + L\delta/\widetilde{\Phi}_m)$, vanishing as the projection sharpens and the cluster radius shrinks. Symmetry holds exactly within a cluster and approximately (with explicit slack) across clusters.
    \item \textbf{$\varepsilon$-Efficiency (Budget Balance).}\\
    \textit{Statement.} The sum of individual valuations equals the federation's total utility up to bounded slack.\\
    \textit{Derivation.} The Group Shapley values satisfy exact efficiency at the meta-agent level: $\sum_{m=1}^M \Phi_m = \nu(\mathcal{C})$ by construction (the inter-group Hoeffding estimator preserves the Shapley axiom of efficiency in expectation). In the allocation phase, the within-group weight normalizer is
    $$\sum_{i \in G_m} w_i \;=\; \frac{\sum_{i\in G_m}\mathbb{I}(s_i{>}0)\exp(s_i/\tau')}{\xi + \sum_{j\in G_m}\mathbb{I}(s_j{>}0)\exp(s_j/\tau')},$$
    where $\xi$ is the numerical-stability term in the denominator (we use $\xi = 10^{-8}$). Two regimes follow from the explicit Budget Re-normalization step in Algorithm~\ref{alg:hgs_shapley}:
    \emph{(i) Active group ($G_m^+ \neq \emptyset$):} the denominator differs from the numerator by exactly $\xi$, so $\sum_{i\in G_m} w_i = 1 - \xi/(Z_m + \xi) = 1 - \mathcal{O}(\xi)$ as $\xi \to 0^{+}$.
    \emph{(ii) Fully adversarial group ($G_m^+ = \emptyset$):} all $w_i = 0$, and the deployed Algorithm~\ref{alg:hgs_shapley} explicitly reabsorbs $\Phi_m$ onto the active groups via $\widetilde{\Phi}_m = \Phi_m\,\Phi_{\text{tot}}/\Phi_{\text{act}}$, so $\sum_{m\in\mathcal{A}} \widetilde{\Phi}_m = \sum_{m=1}^M \Phi_m = \nu(\mathcal{C})$ and this case contributes no efficiency loss.
    Combining, the total allocated utility is $\sum_{i\in\mathcal{C}}\phi_i = \sum_{m\in\mathcal{A}} \widetilde{\Phi}_m(1 - \mathcal{O}(\xi)) = \nu(\mathcal{C}) - \mathcal{O}(\xi)$. Adding the $\mathcal{O}(L\delta)$ centroid-substitution slack from Theorem~\ref{theorem:clustering_error} gives the stated overall efficiency gap of $\mathcal{O}(L\delta + \xi)$, which vanishes in the limit $M \to N$. \qed
\end{enumerate}
\section{Full Experimental Details}
\label{app:impl_details}

\subsection{Dataset Statistics and Partition Strategies}

\textbf{CIFAR-10} ($N{=}16$): 60,000 images across 10 classes. Dirichlet($\alpha{=}0.5$) partition to simulate moderate label heterogeneity. Used exclusively for exact $2^{16}$ Shapley ground-truth validation; federation size kept at $N{=}16$ to make enumeration tractable with caching on a single GPU.

\textbf{FEMNIST} ($N{=}500$): 805,263 character images from 3,550 writers (LEAF benchmark~\cite{Caldas2018LEAF}). Subsampled 500 active clients. Heterogeneity arises from natural feature distribution shifts (writing styles) rather than label skew. Standard 80/10/10 train/val/test split per client.

\textbf{CheXpert} ($N{=}100$): 224,316 chest radiographs from 65,240 patients across 14 pathology labels~\cite{irvin2019chexpert}. Partitioned strictly by Patient ID into 100 institutional silos. 10 clients designated as ``Hero'' clients by holding exclusive access to at least one rare pathology ($<$1\% prevalence).

\textbf{StackOverflow} ($N{=}1000$): 135M tokens from 342,477 users~\cite{stackoverflowdataset2022}. Subsampled 1,000 cross-device clients. Next-word prediction over a 10,000-word BPE vocabulary. GPT-2 Small~\cite{radford2019language} (117M parameters) creates the most challenging projection regime.

\subsection{Implementation Details}

\textbf{Vision tasks} (FEMNIST, CheXpert): ResNet-18~\cite{he2016residual} (FEMNIST) and DenseNet-121~\cite{huang2017densely} (CheXpert) trained with SGD~\cite{sutskever2013importance} (Nesterov momentum $0.9$), initial learning rate $0.01$, Cosine Annealing schedule~\cite{loshchilov2017sgdr}. Preprocessing: random crop, horizontal flip, normalization.

\textbf{NLP task} (StackOverflow): GPT-2 Small~\cite{radford2019language} trained with AdamW~\cite{loshchilov2019decoupled}, learning rate $5{\times}10^{-5}$, weight decay $1{\times}10^{-4}$. Byte-pair encoding, vocabulary limit 10,000.

\textbf{Shared settings}: 5 local epochs per round, batch size 64, label smoothing $0.1$, Dropout $0.2$, 200 communication rounds. Validation set: 10\% of data held by server.

\textbf{HGS-Shapley hyperparameters} (fixed across all domains):
\begin{itemize}\setlength{\itemsep}{-0.5ex}
  \item Projection dimension: $k{=}1024$ (vision), $k{=}2048$ (text)
  \item Cluster count: $M{=}\lfloor\sqrt{N}\rfloor$
  \item Permutation budget: $K{=}50$
  \item Temperature: $\tau{=}0.05$
  \item EMA decay: $\beta{=}0.9$
\end{itemize}

\textbf{Hardware}: 8$\times$NVIDIA H100 80GB (NVLink). Results averaged over 5 seeds (10 for headline numbers). Runtime measured as wall-clock time per federated round on a single H100 (valuation phase only).

\subsection{Baseline Configurations}

\textbf{TMC-Shapley (Oracle)}: 10,000 permutations, truncation threshold $\epsilon{=}0.01{\times}\nu(\mathcal{C})$. Required $>$20,000 GPU-hours across all datasets.

\textbf{GTG-Shapley}~\cite{Liu2022GTGShapley}: Gradient Truncation Gradient method; 50 gradient steps per evaluation, same permutation budget $K{=}50$ for fair comparison.

\textbf{Krum}~\cite{blanchard2017machine} \textbf{/ Trimmed Mean}~\cite{wang2025federated}: $f{=}0.3N$ assumed Byzantine fraction for Krum. Trimmed Mean discards top/bottom 10\% per coordinate.

\textbf{Fed-Influence}~\cite{xue2021toward}: Hessian-vector product approximation of Leave-One-Out cost using Gauss-Newton approximation with damping $\lambda{=}0.01$.

\end{document}